\documentclass[sageh,Afour]{sagej}
\usepackage[T3,T1]{fontenc}
\DeclareSymbolFont{tipa}{T3}{cmr}{m}{n}
\DeclareMathAccent{\invbreve}{\mathalpha}{tipa}{16}

\usepackage{graphicx}
\usepackage[super]{nth} %to type 1st,2nd,3rd,... , use the command \nth {<number>}
\usepackage[normalem]{ulem} % to strike out
\usepackage{color}
\usepackage{xcolor} %for coloring text
\usepackage{siunitx} %e.g. \ang{45} to show 90 degree
\usepackage{textcomp} %for a nice registered trademark, e.g. MATLAB\textsuperscript{\textregistered}
\usepackage{tikz}
\usepackage{xcolor}

\usepackage{pifont}

\usepackage{array}
\usepackage{booktabs}
\usepackage{floatrow}  %needed to push location of table caption above table https://tex.stackexchange.com/questions/22751/how-to-force-table-caption-on-top
\usepackage{multirow}  %needed for multi-row table
\usepackage{tabularx}  %needed for tabularx environment for nicer column alignment in a table
\usepackage{colortbl} %needed for colored cells in Tabularx see http://tex.stackexchange.com/questions/148784/colored-diagonal-divided-table-cell
\usepackage{tabu}
\usepackage{footnote}% placed after tabularx so it does not conflict
\usepackage{tablefootnote}
\usepackage{arydshln} % this one is required for dashed line in table
\makeatletter
\newcommand{\thickhline}{%
    \noalign {\ifnum 0=`}\fi \hrule height 0.8pt
    \futurelet \reserved@a \@xhline
}
\newcolumntype{"}{@{\hskip\tabcolsep\vrule width 2pt\hskip\tabcolsep}}
\newcolumntype{C}[1]{>{\centering\arraybackslash}p{#1}}

\definecolor{Gray}{gray}{0.85}
\newcolumntype{a}[1]{>{\columncolor{Gray}}C{#1}}
\newcolumntype{b}{>{\columncolor{Gray}}c}
\makeatother
\newif\ifTrackChanges   %define a conditional variable called TrackChanges
\TrackChangesfalse %set TrackChanges to be false. %Uncomment to make a clean version (you have to comment the line above)
\definecolor{deep-red}{RGB}{192, 0, 0}
\definecolor{deep-purple}{RGB}{120, 0, 170}
\definecolor{good-green}{RGB}{0,175,0}
\definecolor{purple}{RGB}{210, 0, 210}
\definecolor{alizarin}{rgb}{0.82, 0.1, 0.26}

\usepackage[textsize=scriptsize, bordercolor=white,backgroundcolor=gray!30,linecolor=black,colorinlistoftodos]{todonotes}  %for adding comments. See http://www.texample.net/tikz/examples/todo-notes/.
\usepackage[normalem]{ulem} % added for strikthrough text use   \sout{text you want to strike through
\usepackage{soul} %added for \texthl for the highlight and also for \ul for underline across wrapped text

\usepackage[mathscr]{euscript}
\ifTrackChanges
    \newcommand{\cut}[1]{{\color{lightgray}{#1}}}
    \newcommand{\remind}[2]{{[\colorbox{cyan}{#1}]}{\color{blue}{#2}}}  %reminder to a reviewer about existing text in response to a comment
    \newcommand{\corrlab}[2]{{[\colorbox{yellow}{#1}]~}{\color{red}{#2}}} %correction with a label in response to a reviewer
    \newcommand{\cusst}[1]{{\st{#1}}}  % struck out test 
\else
     \newcommand{\cut}[1]{}
     \newcommand{\remind}[2]{#2}
    \newcommand{\corrlab}[2]{#2}
    
    \newcommand{\cusst}[1]{} 
\fi
\usepackage{algorithm,algorithmicx,algpseudocode}

\algnewcommand\algorithmicinput{\textbf{Input:}}
\algnewcommand\Input{\item[\algorithmicinput]}

\algnewcommand\algorithmicoutput{\textbf{Output:}}
\algnewcommand\Output{\item[\algorithmicoutput]} 

\usepackage{enumerate}
\usepackage{enumitem}
\usepackage{pifont}   %needed for special bullet symbols using \ding
\newcommand{\arrowbullet}{\par \noindent \ding{227}~}

\usepackage{amsmath}
\DeclareMathSymbol{:}{\mathord}{operators}{"3A}  %to reduce space around colons in math mode. See https://tex.stackexchange.com/questions/432778/reduce-spacing-around-colons-in-math-mode. 
\usepackage{yhmath} %note this package will generate type 3 fonts unless you have 1.6 installed
\usepackage{mathtools} % for rotation matrix e.g. \prescript{4}{0}{R} and for summation
\usepackage{amssymb} %for trianleq symbol
\usepackage{xfrac} % needed for diagonal fraction \sfrac{a}{b} command
\usepackage{sidecap}
\usepackage{cancel}
\usepackage{wasysym}  % for diameter symbol e.g. \diameter
\usepackage{mathrsfs} 
\usepackage{tikz}
\usetikzlibrary{matrix,decorations.pathreplacing}
\newcommand{\realfield}[1]{\hbox{I \kern -.5em R}^{#1}}   %needed to
\newcommand {\mb}[1]{\mathbf{#1}}
\newcommand {\bs}[1]{\boldsymbol{#1}}
\newcommand{\uvec}[1]{\hat{\mathbf{#1}}}

\newcommand{\T}{^{\mathrm{T}}}  %shortcut for transpose
\newcommand{\Rot}[2]{{}^{#1}\mb{R}_{#2}}

\newcommand{\eff}{\mathfrak{f}}

\newcommand*\circled[1]
{\tikz[baseline=(char.base)]{\node[circle,minimum size=6pt,draw=black,inner sep=0.25pt](char){\scriptsize #1};}}
\newcommand{\htf}[2]{{^{#1}\mathbf{T}}_{#2}}  %example: \htf{0}{1}

\graphicspath{{figures/pdf}}

\usepackage{fancyhdr}
\fancypagestyle{firstpage}{
  \fancyhf{} % clear header/footer
  \fancyhead[L]{International Journal of Robotics Research. Accepted Version.}
}

\title{A Modal-Space Formulation for Momentum Observer Contact Estimation and Effects of Uncertainty for Continuum Robots}
\date{2024}
\author{Garrison~L.H.~Johnston\affilnum{1},~Neel~Shihora\affilnum{1},~Nabil~Simaan\affilnum{1}}
\affiliation{\affilnum{1}Department of Mechanical Engineering, Vanderbilt University, Nashville TN 37235}
\corrauth{Nabil Simaan}
\email{nabil.simaan@vanderbilt.edu}
\begin{document}
\runninghead{Johnston, G., et al.}

\begin{abstract}
Contact detection for continuum and soft robots has been limited in past works to statics or kinematics-based methods with assumed circular bending curvature or known bending profiles. In this paper, we adapt the generalized momentum observer contact estimation method to continuum robots. This is made possible by leveraging recent results for real-time shape sensing of continuum robots along with a modal-space representation of the robot dynamics. In addition to presenting an approach for estimating the generalized forces due to contact via a momentum observer, we present a constrained optimization method to identify the wrench imparted on the robot during contact. We also present an approach for investigating the effects of unmodeled deviations in the robot's dynamic state on the contact detection method and we validate our algorithm by simulations and experiments. We also compare the performance of the momentum observer to the joint force deviation method, a direct estimation approach using the robot's full dynamic model. \corrlab{R1-1}{We also demonstrate a basic extension of the method to multisegment continuum robots.} Results presented in this work extend dynamic contact detection to the domain of continuum and soft robots and can be used to improve the safety of large-scale continuum robots for human-robot collaboration.     
\end{abstract}  
% \begin{IEEEkeywords}
% Continuum robots, soft robots, collaborative robots, contact detection
% \end{IEEEkeywords}
\keywords{Continuum robots, soft robots, contact detection, collaborative robots}

\maketitle
\thispagestyle{firstpage}  %FOR ARIXV
% section input
\begin{abstract}
Contact detection for continuum and soft robots has been limited in past works to statics or kinematics-based methods with assumed circular bending curvature or known bending profiles. In this paper, we adapt the generalized momentum observer contact estimation method to continuum robots. This is made possible by leveraging recent results for real-time shape sensing of continuum robots along with a modal-space representation of the robot dynamics. In addition to presenting an approach for estimating the generalized forces due to contact via a momentum observer, we present a constrained optimization method to identify the wrench imparted on the robot during contact. We also present an approach for investigating the effects of unmodeled deviations in the robot's dynamic state on the contact detection method and we validate our algorithm by simulations and experiments. We also compare the performance of the momentum observer to the joint force deviation method, a direct estimation approach using the robot's full dynamic model. \corrlab{R1-1}{We also demonstrate a basic extension of the method to multisegment continuum robots.} Results presented in this work extend dynamic contact detection to the domain of continuum and soft robots and can be used to improve the safety of large-scale continuum robots for human-robot collaboration.     
\end{abstract}  
% \begin{IEEEkeywords}
% Continuum robots, soft robots, collaborative robots, contact detection
% \end{IEEEkeywords}
\keywords{Continuum robots, soft robots, contact detection, collaborative robots}
\section{Introduction} \label{sec:intro}
\par
\begin{figure}
  \centering
  \includegraphics[width=0.9\columnwidth]{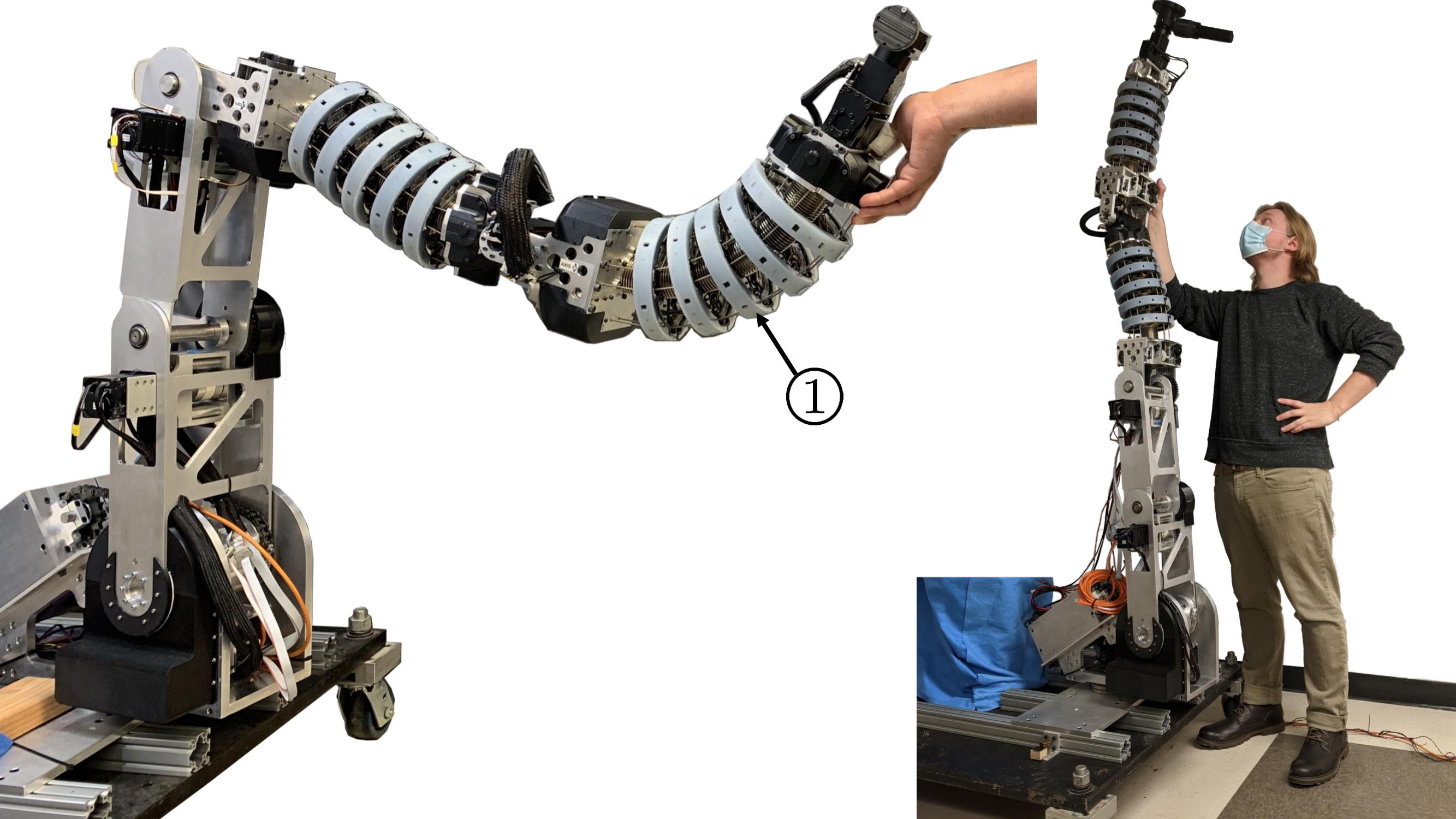}
  \caption{A collaborative robot for cooperative manipulation in confined spaces using continuum segments. Contact detection and wrench estimation is a key component of ensuring worker safety during human-robot interaction. \protect\circled{1} Robot sensory skin as described in \cite{Abah2022_sensor_disk}}\label{fig:Vanderbot}
\end{figure}

 Human-robot collaboration presents a host of sensing and situational awareness challenges in order to ensure worker safety. These challenges are exacerbated for \textit{in-situ} collaborative robots (ISCRs) where robots must safely coexist and interact with a human co-worker within a confined space. An example of such an ISCR is shown in Fig.~\ref{fig:Vanderbot}. In \cite{Abah2022_sensor_disk} we presented our work on achieving active safety using sensory disks \circled{1} along the length of the continuum robot. While these sensory disks sense proximity, contact, and contact force along the length of the robot, in this paper, we consider an alternate approach for contact estimation using the robot's dynamics because we realize that achieving enhanced active safety requires several \textit{independent} methods to ensure fail-safe operation. The scope of this paper is therefore focused on identifying the challenges of contact detection and wrench estimation for continuum robots and on presenting what we believe is the first adaptation of the generalized momentum observer (GMO) method for continuum robots. 
\corrlab{R3-1.1}{\section{Related Works}
\subsection{General Contact Estimation}}
\par In past works, to address the challenge of contact detection and estimation, other researchers have also incorporated sensors into robots in order to estimate the location and magnitude of the wrench applied during contact. Examples of this include sensory skins such as \cite{Yamada2005_sensor}, \cite{Abah2019_sensor_disk}, and \cite{Abah2022_sensor_disk}, strain gauges placed onto the body of the robot as in \cite{Garcia2003_sensor} and \cite{Malzahn2014_sensor}, force/torque sensors in the robot's wrist and/or base as in  \cite{Lu2005_sensor} and \cite{Shimachi2008_sensor}, vibration and acoustic analysis as in \cite{Richmond2000_acoustic}, \cite{Min2019_vibrations}, and \cite{Fan2020_sensor}, and joint torque sensors as in \cite{Popov2017_sensor_neural_net}. 
%--------------------
%--------------------
\par Sensor-based methods can suffer from dead zones where sensors are not present and can be cost-ineffective. Therefore, researchers explored methods for contact detection that do not rely on adding extra sensors to the robot. The earliest technique is the joint force/torque deviation (JFD) method given in \cite{Takakura1989_observers}, \cite{Suita1995_JFD}, \cite{Yamada1996_JFD} and \cite{Bajo2010_continuum_collision}. This method is also referred to as \textit{direct estimation} by some authors (e.g., in \cite{haddadin2017_collsion_review}). The JFD method consists of subtracting model-predicted generalized forces from the sensed generalized forces (e.g., using joint torque sensing). The model-predicted generalized forces may be from a full dynamic model as in \cite{Suita1995_JFD} or from a statics model as in \cite{Bajo2010_continuum_collision}. A model of the joint level friction is sometimes added to increase the accuracy of the method. The downside of this method is that it relies on joint-level current (or torque) sensing and is highly sensitive to the accuracy of the dynamic/static model. This can lead to false positives due to modeling errors and/or unmodeled friction. For methods that rely on the full dynamic model to predict the generalized forces such as \cite{Takakura1989_observers}, \cite{Suita1995_JFD}, and \cite{Yamada1996_JFD}, a numerical second-derivative of the joint positions is required which can lead to extremely noisy estimation signals. Additionally, JFD works best for slow motions where viscous friction and the effect of unmodeled dynamics are small.
\par Due to the aforementioned issues with estimation noise, it may be desirable to use methods that do not rely on the second derivative of the joint position and provide estimation noise filtering. These methods are based on the fault detection literature such as \cite{Wunnenberg1990_fault_detection}, \cite{Freyermuth1991_fault_detection}, \cite{Caccavale1997_observer}, \cite{Hammouri1999_nonlinear_observer}, and \cite{DePersis2001_geometric_nonlinear_fault_detection} and typically rely on knowledge of the robot's dynamic parameters. They work by observing an unexpected deviation in some aspect of the robot's dynamic state such as its energy, joint velocity, or generalized momentum. A detailed review of these methods is given in \cite{haddadin2017_collsion_review}. The most common of these methods is the generalized momentum observer (GMO). This method was first developed in \cite{DeLuca2003_GMO} and has been used in a wide range of applications such as flying robots in \cite{Tomic2017_GMO}, humanoids in \cite{Vorndamme2017_GMO}, redundant manipulators in \cite{DeLuca2008_GMO} and \cite{Cho2012_GMO}, and robots with series-elastic actuators in  \cite{DeLuca2006_detection_elastic} and \cite{Kim2015_GMO_flexible_joints}. The GMO has also been combined with Kalman filters in \cite{Wahrburg2015_GMO} and \cite{Wahrburg2018_GMO}, additional sensory input in \cite{Buondonno2016_GMO}, and friction models in \cite{Lee2015_GMO} and \cite{Lee2016_GMO} to improve the performance of the observer. The performance of the observer under characterized dynamic model uncertainty has also been studied in \cite{Briquet-Kerestedjian2017_GMO_uncertanty} and \cite{Li2021_GMO_uncertanty}.
\corrlab{R2-4}{It should also be noted that the GMO bares some similarities to the time-delay control paradigm introduced in \cite{Youcef-Toumi1988_time_delay_control} as it relies on previous measurements of the system state to estimate the contributions of external disturbances. }
\par The GMO method for contact estimation has been used predominantly for rigid-link robots with/without series-elastic actuators. A detailed review of contact detection and localization methods for continuum robots is presented next in the related works section. Despite its successful use for various robot architectures, adapting the GMO method has not been previously applied to continuum robots. Furthermore, investigating the effects of state uncertainty on the performance of the contact estimator of a continuum robot has not been carried out. 
\par The contribution of this paper is in presenting a modal space formulation of the GMO and verification of the GMO method to \textit{variable curvature} continuum robots while taking into account uncertainty in the dynamic state of the robot. Unlike serial rigid-link robots, the shape of continuum robots is heavily dependent on external loading. To enable successful adaptation of the GMO method for continuum robots, one must overcome the challenge due to the load-induced shape changes. One, therefore, needs a modeling framework that takes this information into account. For these reasons, we leverage our recent result on real-time shape sensing in \cite{Orekhov2023_shape_sensing} along with a dynamic formulation in the modal curvature space to present a formulation of the GMO method that can be used for continuum and soft robots. Relative to prior works in GMO-based contact estimation, we present a unique constrained optimization formulation for solving the problem of contact wrench estimation. We also present a model of the effects of dynamic state uncertainty on the output of the GMO and leverage this result to make recommendations for setting detection thresholds. Lastly, we compare the performance of the GMO to the performance of the JFD method in both simulation and experiment.  
\par The next section presents related works while focusing on the most relevant works on contact detection/estimation for continuum robots. The following sections present the shape sensing approach and a dynamic formulation in the space of curvature modal factors. We then present an adaptation of the GMO method and an investigation of state parameter uncertainty on expected performance. Finally, sections \ref{sec:simulations} and \ref{sec:experiments} present a simulation-based study and experimental validation on the isolated distal continuum segment of the robot shown in Fig.~\ref{fig:Vanderbot}. \corrlab{R1-1}{Lastly, in section \ref{sec:multiseg} we present a preliminary extension of the method to multisegment continuum robots.}

\corrlab{R3-1.2}{\subsection{Contact Estimation for Continuum Robots}}
In addition to the seminal works on collision detection reviewed above, the most relevant work to this paper is \cite{DeLuca2003_GMO}, which we build on and extend in this paper. Within the scope of contact detection/estimation for continuum robots, the most relevant works are \cite{Xu2008_continuum_collision}, \cite{Bajo2010_continuum_collision}, \cite{Bajo2012TRO_continuum_collision}, and \cite{Santina2020_continuum_disturbance_observer}. In \cite{Bajo2010_continuum_collision} the authors considered the JFD method and the effects of bounds of joint-force sensing uncertainty on the detectability of contact. In \cite{Bajo2012TRO_continuum_collision}, a kinematics-based contact detection and localization approach based on motion tracking and detection in the shift of the instantaneous screws of motion was presented and shown to be quite effective for continuum robots that bend in circular shapes. \cite{Roy2016_dynamic_effect_collision} simulated the effects of dynamics on this method assuming constant curvature bending profiles. In \cite{Rucker2011_continuum_collision}, the authors combine Cosserat rod theory and extended Kalman filters to estimate the applied wrench to the tip of a continuum robot in quasi-static configurations. More recently, in \cite{Ashwin2021_continuum_collision_detection}, the authors presented contact detection and estimation for wire-actuated continuum robots subject to quasi-static modeling assumptions. \corrlab{R2-1.1}{Lastly, in \cite{Leung2024_continuum_collision_detection}, the authors present two tip-force estimation techniques for a two segment continuum robot using constant curvature assumptions. Their first technique uses a cantilever beam model to estimate the tip force and their second technique uses support vector regression to estimate the difference between no-load tendon forces and loaded tendon forces.}  

\par Table~\ref{tab:relevant_works_compare} summarizes the key modeling assumptions and capabilities supported by this work compared to the most relevant prior works. The limitations of these prior works on continuum robot contact estimation/detection are:
\begin{itemize}[label=\ding{227},noitemsep,leftmargin=!, rightmargin=0.0cm,labelindent=0.3cm,itemindent=0.0cm]
\item The assumption of a known external load state (mostly unloaded) and known bending shape profiles of the continuum robot segments (e.g., most works in Table~\ref{tab:relevant_works_compare} assume constant curvature or use piece-wise constant curvature (PCC)). For example, \cite{Xu2008_continuum_collision} assumed constant curvature and a force at the segment tip while \cite{Bajo2012TRO_continuum_collision} assumed constant curvature and detected contact along the length of a segment. Finally, \cite{Chen2021_force_estimation_soft} assumed a planar curvature experimentally captured via a modal representation of the local tangent angle of a fiber-reinforced bellow actuator acting as a single-segment soft robot.
\item The methods do not estimate the magnitude of the contact force and assume negligible dynamics. For example, \cite{Bajo2012TRO_continuum_collision} assumed negligible dynamics in their kinematics-based contact detection and localization approach. In \cite{Santina2020_continuum_disturbance_observer}, the authors estimated contact forces in soft robots while assuming PCC, ignoring Coriolis/centrifugal and inertial effects, and assuming extrinsic sensing of the robot shape using optical tracking.
\end{itemize}
\begin{table*}[]
\caption{Summary of the continuum robot contact estimation literature}
\label{tab:relevant_works_compare}
\small
\begin{tabular}{|c|C{1.5cm}|C{1.9cm}|C{1.7cm}|C{1.7cm}|C{2.1cm}|}
\hline
\rule{0pt}{2ex}\textbf{Method} & \textbf{Dynamics Included?} & \textbf{Contact \quad Localization?} & \textbf{Wrench \quad Estimation?} & \textbf{General Curvature?} & \textbf{Intrinsic Shape Sensing?} \\\hline
\cite{Xu2008_continuum_collision} &  &  & \checkmark &  & \\ \hline
\cite{Bajo2010_continuum_collision} &  & \checkmark &  &  & \\ \hline
\cite{Bajo2012TRO_continuum_collision}&  & \checkmark &  & &  \\ \hline
\cite{Roy2016_dynamic_effect_collision}& \checkmark & \checkmark &  & & (Sim. only) \\ \hline
\cite{Rucker2011_continuum_collision}&  &  & \checkmark & (PCC) & (Sim. only) \\ \hline \cite{Santina2020_continuum_disturbance_observer}&  &  & \checkmark & (PCC) & \\ \hline
\cite{Chen2021_force_estimation_soft}&  & \centering\checkmark &  &  \checkmark & \\ \hline
\cite{Ashwin2021_continuum_collision_detection}&  & \checkmark & \checkmark &  &  \\ \hline
\cite{Zeng2023_continuum_koopman} & & & \checkmark & \checkmark & (Sim. only)\\ \hline
\corrlab{R2-1.2}{\cite{Leung2024_continuum_collision_detection}} & & & \checkmark &  &\\ \hline
\textbf{This work} & \checkmark &  & \checkmark & \checkmark & \checkmark \\ \hline
\end{tabular}
\end{table*}
\par While these approaches can work very well for small continuum robots with negligible dynamics and gravitation-induced deflections, large robots such as in Fig.~\ref{fig:Vanderbot} require special consideration of their dynamics and the ensuing shape deflections.
\par The goal of this work is to address the limitations of prior works as shown in Table~\ref{tab:relevant_works_compare} via an extension/adaptation of the GMO method to continuum robots subject to dynamic loading and with general bending shape profiles. The robot shown in Fig.~\ref{fig:Vanderbot} is very large and therefore its segments undergo significant passive deflections violating the constant curvature assumptions. This robot is also designed to work within confined spaces where external optical tracking may not be possible, therefore a combined approach achieving intrinsic shape sensing, contact detection, and contact wrench estimation is sought in this work.
\par Another set of key works that we build on and extend is the literature on continuum robot dynamics and kinematics. In this work, we build on Chirikjian's \cite{Chirikjian1994_modal,Chirikjian1995_modal,Chirikjian1994_hyperredundant_dynamics} early works on modal space representation of hyperredundant robots. We present a modal-space dynamics model for the purpose of formulating the GMO contact estimation approach and analyzing the effects of uncertainty. %\todo{I do not think the dynamics review is needed here}%Although we don't claim to make a contribution to the dynamic modeling of continuum robots, dynamic modeling of continuum robots is a necessary step to implementing the GMO for continuum robots. For this purpose, we will present a brief literature review of dynamic models for continuum robots. This literature review will help select a model to be used for the proposed work. There have been many techniques used to generate dynamic models of continuum robots presented in the literature. Perhaps the most common is the Euler-Lagrange method \cite{Book1984_recursive_lagrangian,Tatlicioglu2007_extensible_continuum_dynamics,Tatlicioglu2007_planar_continuum_dynamics,Godage2011_shape_function_variable_length_dynamics,He2013_lagrange,Falkenhahn2014_cc_lagrange,Roy2016_dynamic_effect_collision}. In \cite{Chirikjian1994_hyperredundant_dynamics}, Chirikjian used continuum mechanics to derive the equations of motion of curvilinear snake robots. Other dynamic modeling techniques that have been used in the literature include the recursive Newton-Euler method \cite{Khalil2005_recursive_newton_euler,Kang2011_recursive_newton_euler}, Hamilton's principle \cite{Gravagne2003_large_deflection_continuum}, Kane's method \cite{Everett1989_kanes_method}, lumped parameter models \cite{Jung2011_lumped_parameter}, virtual power modeling \cite{Rone2014_virtual_power}, Cosserat rod theory \cite{Till2019_cosserat_dynamics}, pseudo-rigid body models \cite{Mochiyama2003_flexible_dynamics}, and  mass-spring-damper networks \cite{Guo2015_continuum_dynamics}.
\section{Problem Definition \& Modeling Assumptions}
\par This paper focuses on presenting an extension of the GMO contact estimation method to continuum robots. Because these robots are very different from traditional serial rigid-link robots, these robots have significant sources of geometric, dynamic, and mechanics uncertainty. This combination of challenges requires solutions for online shape sensing, updated kinematics, and updated dynamics in a way that allows reliable application of the GMO algorithm. Therefore, the two problems targeted by this paper are:
\arrowbullet Collision Estimation: detect contact and estimate the wrench applied to the robot during contact.
\arrowbullet Effects of Uncertainty: formulate the effects of uncertainty and provide an estimate of the effects of uncertainty on the GMO estimator, thereby delineating the limits of contact detectability due to uncertainty. 
\par \corrlab{R2-2}{In dynamics model presented in this paper, we make the following assumptions:
\begin{itemize}
[label=\ding{227},noitemsep,leftmargin=!, rightmargin=0.0cm,labelindent=0.3cm,itemindent=0.0cm]
    \item The friction in the actuation tendons can be modeled as concentrated forces at the contact points between the actuation tendons and the robot body.
    \item The friction can be modeled as a smooth Coulomb friction model
    \item Viscous effects that may arise due to gearhead losses can be neglected
    \item The robot is torsionally rigid. This assumption is justified since the robot shown in Fig. \ref{fig:Vanderbot} has bellows with very high torsional stiffness as reported in \cite{Orekhov2023_shape_sensing}.
    \item The central backbone of the robot includes a superelastic nickel-titanium (NiTi) rod. NiTi is known to exhibit a nonlinear, hysteretic stress-strain curve \cite{Liu1999_nitinol}. Because we mostly operate in the second stress-strain curve plateau that occurs past approximately 2\% strain \cite{Liu1999_nitinol}, we neglect these effects and assume a linear elasticity model for simplicity.
    \item \corrlab{R1-5}{Although the method can handle more complex contact types, we evaluate the method in simulation and experiment using only point contacts. }
\end{itemize}}
\cut{Our modeling assumptions include the assumption of concentrated friction forces at contact points between the actuation tendons and the robot. We assume a smooth Coulomb friction model while neglecting viscous effects that may arise due to gearhead losses. We also assume that the continuum robot is torsionally rigid. This assumption is justified since the robot shown in Fig. \ref{fig:Vanderbot} has bellows with very high torsional stiffness as reported in \cite{Orekhov2023_shape_sensing}. We also assume that the mass of the tendons is negligible compared to the mass and inertia of the continuum robot shown in Fig.~\ref{fig:Vanderbot}. Additionally, the central backbone of the robot includes a superelastic nickel-titanium (NiTi) rod. NiTi is known to exhibit a nonlinear, hysteretic stress-strain curve \cite{Liu1999_nitinol}. Because we mostly operate in the second stress-strain curve plateau that occurs past approximately 2\% strain \cite{Liu1999_nitinol}, we neglect these effects and assume a linear elasticity model for simplicity.  Lastly, although the method can generalize to multi-segment continuum robots, in this paper we will limit ourselves to single-segment robots.}

\section{Modal Kinematics}
\remind{R1-3.1}{The instantaneous shape of the continuum segment shown in Fig. \ref{fig:shape_sensing} will be represented using a modal representation following the model presented in \cite{Orekhov2023_shape_sensing}. For completeness, we briefly summarize the modal kinematics below while using the nomenclature presented in Table~\ref{tab:nomenclature}.}
\par The continuum segment of Fig. \ref{fig:shape_sensing} uses two capstans with two fixed-length wire loops to bend in any direction. The robot uses torsionally rigid metal bellows assembled on a superelastic NiTi rod. Since the bellows are torsionally very stiff, we assume the robot does not experience torsional strains and that the robot has a fixed arc length $L$. 
%%%%%%%%%% NOMENCLATURE TABLE %%%%%%%%%%
\bgroup
\def\arraystretch{1.5}%
\begin{table}[h]
	\centering
	\footnotesize
	\caption{Nomenclature}
	\begin{tabular}{m{0.12\columnwidth} m{.8\columnwidth}}
		\hline
		\textbf{Symbol} & \textbf{Description} \\
		\hline
		\{F\}
		& Designates a right-handed frame with unit vectors $\uvec{x}_f,\uvec{y}_f,\uvec{z}_f$ and $\mb{f}$ as its origin.\\
		\hdashline
        \{T(s)\} &  local frame at segment arc length $s$. It has its $\uvec{z}_t$ axis tangent to the central backbone of the continuum robot.\\
		\hdashline
        $\Rot{a}{b}$ & A rotation matrix describing the orientation of frame \{B\} relative to frame \{A\}.\\[2pt]
        \hdashline
        $(\cdot)^\prime$ &  is a shorthand notation for $\frac{d(\cdot)}{ds}$.\\
        \hdashline
        $\dot{(\cdot)}$ &  is a shorthand notation for $\frac{d(\cdot)}{dt}$, i.e. time derivative.\\
        \hdashline
          $\mb{x}^\wedge$ & 
          The exterior product matrix representation of a vector. 
           \begin{equation*}
                 \mb{x}^\wedge=\begin{cases}
			         \begin{bmatrix} 
                        0 & -x_3 & x_2 \\
                        x_3 & 0 & -x_1\\
                        -x_2 & x_1 & 0
                        \end{bmatrix} \in so(3) & \textrm{for~} \mb{x}=\begin{bmatrix}
                            x_1 \\ x_2 \\x_3
                        \end{bmatrix}\\
                      \begin{bmatrix}
                        \bs{\omega}^\wedge & \mb{v} \\ 
                        \mb{0} & 0
                      \end{bmatrix} \in se(3)  &
                      \mb{x}=\begin{bmatrix}\mb{v}\in\realfield{3} \\ \bs{\omega}\in\realfield{3} \end{bmatrix}              
		      \end{cases}
            \end{equation*}    
        \\	
    \hdashline
    $\left(\mb{X}\right)^\vee$ & 
    An inverse operator of $(\cdot)^\wedge$ such that:
    \begin{equation*}
        \mb{X}^\vee=\begin{cases}
			[x_1, x_2, x_3]\T, & \textrm{for~} \mb{X}\in so(3)\\
                [\mb{v}\T, \bs{\omega}\T]\T, & \textrm{for~} \mb{X}^\wedge = \begin{bmatrix}\bs{\omega}^\wedge & \mb{v} \\ \mb{0} & 0\end{bmatrix} \in se(3)
		 \end{cases}
   \end{equation*}    
    \\
    \hdashline
    $\mb{I}_n$ & an $n\times n$ identity matrix\\
    \hdashline
		$\bs{\kappa}$ & A generalized force\\
		\hline
    \end{tabular}
    \label{tab:nomenclature}
\end{table}
\egroup
%%%%%%%%%%%%%%%%%%%%%%%%%%%%%%%%%%%%%%%%%%%%%
For a given arc length $s\in[0,L]$ along a continuum segment, a local frame $\mb{T}(s)$ is assigned with its $z$-axis tangent to the backbone and following the increment direction of the arc-length (from the base to the segment tip):
\begin{equation}\label{eq:local_frame}
    \htf{0}{t}(s) = \begin{bmatrix}
                 \Rot{0}{t}(s) & {}^0\mb{p}(s) \\
                 \mb{0}        & 1
    \end{bmatrix}\in SE(3)
\end{equation}
\par As the robot undergoes a deformation parameterized in the local frame as a twist ${}^t\bs{\eta}(s)$,  the local frame described by ${}^0\mb{T}_t(s)$ can be found using the following differential equation: 
\begin{equation}\label{eq:local_frame_diffeq}
    \htf{0}{t}'(s) = \htf{0}{t}(s){}^t\bs{\eta}^\wedge(s)
\end{equation}
\par The body twist ${}^t\bs{\eta}(s)$ can be expressed using the local curvature of the backbone $\mb{u}(s)$: 
\begin{equation}
    {}^t\bs{\eta}^\wedge(s) = \begin{bmatrix}
                             \mb{u}^\wedge(s) & \mb{e}_3 \\
                             \mb{0}              & 0
                             \end{bmatrix} \in se(3)
\end{equation}
where $\mb{e}_3 = [0,\,0,\,1]\T$. This differential equation can be solved using either numerical integration or the Magnus expansion as in \cite{Orekhov2020_magnus}. For the experiments and simulations presented herein, we used the $4^{th}$ order Magnus expansion. For more background on the Magnus Expansion, see \cite{Magnus1954_original_paper,Blanes2009_magnus}.
\par In this work, we choose to represent $\mb{u}(s)$ using a weighted sum of polynomial basis functions. These polynomial functions $\bs{\phi}_x(s)$ and $\bs{\phi}_y(s)$ represent the curvature in the $x$ and $y$ directions, respectively. Because the robot is assumed to be perfectly rigid in torsion, there is zero curvature in the $z$ direction (i.e. $\bs{\phi}_z(s) = \mb{0}$). Using these polynomial basis functions and the scalar weights $\mb{c}_x,\mb{c}_y\in\realfield{3}$, the robot's curvature distribution $\mb{u}(s)$ can be written as: 
\begin{equation}\label{eq:curvature}
    \mb{u}(s) = \begin{bmatrix}
                \bs{\phi}_x\T(s) & \mb{0} \\ 
                \mb{0}                & \bs{\phi}_y\T(s) \\
                \mb{0}                & \mb{0}
     \end{bmatrix}\begin{bmatrix}
     \mb{c}_x \\
     \mb{c}_y
     \end{bmatrix} = \bs{\Phi}(s)\mb{c}
\end{equation}
We chose these polynomial basis functions to be the first three Chebyshev polynomials of the first kind:
\begin{equation}
    \bs{\phi}_x\T(s) = \bs{\phi}_y\T(s) = \left[1,~\frac{2s-L}{L},~\frac{8s^2}{L^2}-\frac{8s}{L}+1\right]
\end{equation}
\remind{R1-4.1}{Chebyshev polynomials of the first kind were chosen over a simple monomial basis because Chebyshev polynomials are bounded to within $\pm1$ and therefore improve scaling of the modal coefficients. Additionally, monomial bases are known to suffer from poor conditioning at higher powers as shown in \cite{Gautschi1975_vandermonde}.} \corrlab{R1-4.2}{For a discussion on how the choice of Chebyshev polynomial order affects the accuracy of the kinematics model see \cite{Orekhov2020_magnus}.}
%----------------------------------------------------------------------------
\subsection{Shape Sensing}
%--------------------
\begin{figure}
  \centering
  \includegraphics[width=0.95\columnwidth]{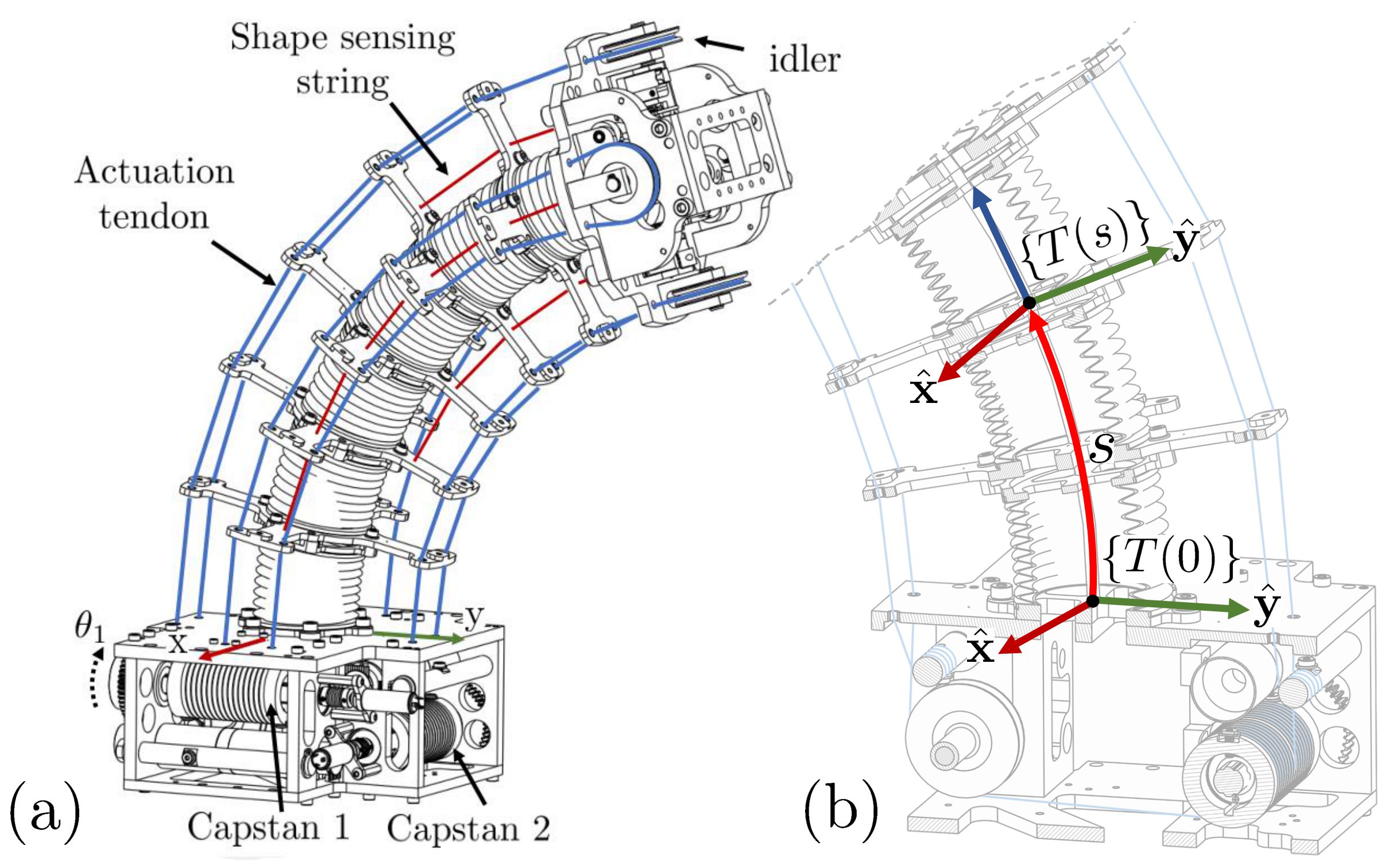}
  \caption{(a) Actuation tendons and passive shape sensing strings whose lengths are used to calculate $\mb{c}$. The figure is reproduced with permission from \cite{Orekhov2023_shape_sensing}.(b) Cross section of the robot showing the backbone arc length coordinate $s$, the local backbone frame $\{T(s)\}$, and the robot base frame $\{T(0)\}$.}\label{fig:shape_sensing}
\end{figure}
%--------------------
As shown in Fig. \ref{fig:shape_sensing}, the robot used in this work is endowed with four string encoders that measure the change in arc length between the end-disk and one of the spacer disks. \corrlab{R2-3.1}{As reported in \cite{Orekhov2023_shape_sensing}, these string encoders are custom-made using Renishaw AM4096 12 b incremental rotary encoders. These custom-made string encoders have an average measurement error of 0.1 mm (0.08\% of the total stroke) and a maximum error of 0.27 mm (0.23\% of total stroke)} \cut{(for more information see \cite{Orekhov2020_workshop})}. Using the string encoder measurements, as well as the change in backbone lengths, we can calculate the modal coefficients $\mb{c}$ that describe the curvature distribution $\mb{u}(s)$. To do this, we define the augmented space of tendon and sensory string extensions relative to the straight home configuration, i.e., $\Delta \bs{\ell}\triangleq \left[\Delta \ell_1, \cdots, \Delta \ell_4, \Delta\ell_{q_1},~\Delta\ell_{q_2} \right]\T$. We then follow the derivation in \cite{Orekhov2023_shape_sensing} to get the matrix $\mb{J}_{\ell c}\in\realfield{6\times6}$ that maps the modal coefficients $\mb{c}$ to $\Delta \bs{\ell}$:  
\begin{equation}\label{eq:J_lc}
    \Delta\bs{\ell} =  \mb{J}_{\ell c}\mb{c}
\end{equation}
Because the robot used in this paper is torsionally stiff and has constant pitch-radius string encoder/actuation tendon routing, $\mb{J}_{\ell c}$ is a constant, square, full-rank matrix \cite{Orekhov2023_shape_sensing} and can therefore be inverted to solve for $\mb{c}$ given measurements of $\Delta\bs{\ell}$.
%----------------------------------------------------------------------------
\subsection{Modal Body Jacobian}
In this section, we will derive the Jacobian $\mb{J}_{\xi c}(s)\in\realfield{6\times6}$ that relates the temporal rates of the modal coefficients $\dot{\mb{c}}\in\realfield{6}$ to the body twist coordinates such that ${}^t\bs{\xi}(s)=\mb{J}_{\xi c}(s)\dot{\mb{c}}$.
\par The body twist of the end-effector can be found using:
\begin{equation}
  {}^t\bs{\xi}^\wedge(s) = \htf{0}{t}^{-1}(s)~{}^0\dot{\mb{T}}_t(s)\in se(3)
\end{equation}
After integration of \eqref{eq:local_frame_diffeq}, ${}^0\mb{T}_t$ becomes a function of curvature $\mb{c}$, therefore, using the chain rule results in:
\begin{equation}
  {}^t\bs{\xi}^\wedge(s) = \sum_{i=1}^{6}\left[\htf{0}{t}^{-1}(s)\left(\frac{\partial}{\partial c_i}\htf{0}{t}(s)\dot{c}_i\right)\right]
\end{equation}
where $c_i$ is the $i^\text{th}$ modal coefficient. The $i^\text{th}$ column of $\mb{J}_{\xi c}$ can be found using.
\begin{equation}\label{eq:J_xic}
  \mb{J}^{[i]}_{\xi c}(s) = \left[\htf{0}{t}^{-1}(s)\left(\frac{\partial}{\partial c_i}\htf{0}{t}(s)\right)\right]^\vee
\end{equation}
To calculate \eqref{eq:J_xic}, we must compute the partial derivative of $\htf{0}{t}(s)$ with respect to $c_i$. Since $\htf{0}{t}(s)$ was computed using the Magnus expansion, $\frac{\partial}{\partial c_i}\htf{0}{t}(s)$ is computed using the derivative of the matrix exponential (see \cite{selig_geometric_2005} and \cite{Orekhov2023_shape_sensing}). 

\subsection{Modal Capstan Jacobian}
In this section, we will derive the Jacobian $\mb{J}_{cq}(s)\in\realfield{6\times2}$ that relates the angular velocity of the capstans $\dot{\mb{q}}\in\realfield{2}$ to the velocity of the modal coefficients $\dot{\mb{c}}\in\realfield{6}$. The tendon speeds $\dot{\bs{\ell}}_q = [\dot{\ell}_{q_1}, ~\dot{\ell}_{q_2}]\T$ can be calculated from $\dot{\mb{q}}$ using:
\begin{equation}
    \dot{\bs{\ell}}_q = \frac{1}{2\pi}\sqrt{(2\pi r_c)^2+\gamma^2}\,\dot{\mb{q}}
\end{equation}
where $r_c$ is the capstan radius and $\gamma$ is the capstan lead. Using $\mb{J}_{\ell c}$, we can write the velocity of the modal coefficients that result from the angular velocity of the capstans:
\begin{equation}\label{eq:J_cq}
    \dot{\mb{q}} = \underbrace{2\pi\left(\sqrt{(2\pi r_c)^2+\gamma^2}\right)^{-1}\begin{bmatrix}\mb{0} & \mb{I}_2\end{bmatrix}\mb{J}_{\ell c}}_{\mb{J}_{qc}}\dot{\mb{c}}
\end{equation}
where the matrix $[\mb{0}, \mb{I}_2]$ accounts for the fact that only the last two values of $\dot{\bs{\ell}}$ correspond to the control tendon lengths.
\section{Continuum Robot Dynamics\\ in the Space of Modal Coefficients}\label{sec:dynamics}
Since the GMO method requires the robot's dynamics, the dynamics of the continuum segment shown in Fig. \ref{fig:shape_sensing} is derived in the space of modal coefficients using the Euler-Lagrange method. The closest work to this section is the work of \cite{Roy2016_dynamic_effect_collision}. However, the dynamic model presented in \cite{Roy2016_dynamic_effect_collision} assumes constant curvature (i.e., circular) bending. In this section, we will relax this assumption. \corrlab{R1-3}{The derivations provided in sections \ref{sec:T} and \ref{sec:V} are extended to general curvature from the constant-curvature models presented in \cite{Roy2016_dynamic_effect_collision}. The friction model presented in section \ref{sec:friction} is unique to this paper.}  
\subsection{Kinetic Energy}\label{sec:T}
The total kinetic energy of the robot can be modeled as the sum of the central backbone's kinetic energy $T_{CB}$, the kinetic energy of the spacer disks $T_{SD}$, and the kinetic energy of the actuation capstans and motors $T_A$.
\begin{equation}
    T = T_{CB}+T_{SD}+T_A
\end{equation}
The models presented in sections \ref{sec:T_CB} and \ref{sec:T_SD} are adapted to general curvature from the constant curvature models presented in \cite{Roy2016_dynamic_effect_collision}. However, the derivations provided in section \ref{sec:T_A} are unique to this paper.
\subsubsection{Central Backbone}\label{sec:T_CB}
The kinetic energy of the central backbone can be found using the robot body twist coordinates ${}^t\bs{\xi}(s)$, the mass per unit length of the backbone $\rho$, and the backbone radius $r$:
\begin{equation}
    T_{CB} = \frac{1}{2}\int_0^L{}^t\bs{\xi}\T(s)\begin{bmatrix}\rho\mb{I} & \mb{0} \\ \mb{0} & \bs{\mathcal{I}}_{CB}\end{bmatrix}{}^t\bs{\xi}(s)ds
\end{equation}
In this equation, $\bs{\mathcal{I}}_{CB}$ is the inertia tensor of a thin, circular cross-section of the backbone:
\begin{equation}
    \bs{\mathcal{I}}_{CB} = \begin{bmatrix} 
                        \frac{1}{4}\rho r^2 & 0                    & 0 \\
                        0                   & \frac{1}{4}\rho r^2  & 0 \\
                        0                   & 0                    & \frac{1}{2}\rho r^2
                     \end{bmatrix}
\end{equation}
Using the modal Jacobian matrix $\mb{J}_{\xi c}(s)$, we can write $T_{CB}$ as a function of the modal coefficient velocities $\dot{\mb{c}}$:
\begin{equation}
    T_{CB} = \frac{1}{2}\dot{\mb{c}}\T\underbrace{\int_0^L\mb{J}_{\xi c}\T(s)\begin{bmatrix}\rho\mb{I} & \mb{0} \\ \mb{0} & \bs{\mathcal{I}}_{CB}\end{bmatrix}\mb{J}_{\xi c}(s)ds}_{\mb{M}_{CB}}\dot{\mb{c}}
\end{equation}
\begin{equation}\label{eq:T_CB}
    T_{CB} = \frac{1}{2}\dot{\mb{c}}\T\mb{M}_{CB}\dot{\mb{c}}
\end{equation}
\subsubsection{Spacer Disks}\label{sec:T_SD}
In the following, we define for each spacer disk the local backbone frame \{T$(s_{d_i})$\}, $i\in[1,\dots,n_d]$ where $n_d$ denotes the number of spacer disks. Since each spacer disk may have its center of mass offset from \{T$(s_{d_i})$\}, we define a center-of-mass body frame \{U$(s_{d_i})$\} having its origin at the center of mass of the $i^{th}$ disk and parallel to \{T$(s_{d_i})$\}. Using ${}^t\mb{p}_{cm_i}\in\realfield{3}$ to denote the location of the disk center of mass in frame \{T$(s_{d_i})$\}, we obtain the body twist of the $i^\text{th}$ spacer disk in a center of mass frame \{U$(s_{d_i})$\} using the adjoint twist transformation:
\begin{equation}\label{eq:disk_twist_in_u}
    {}^u\bs{\xi}_{cm_i} = \underbrace{\begin{bmatrix}
       \mb{I} &  -{}^t\mb{p}^\wedge_{cm_i} \\
       \mb{0} & \mb{I}
    \end{bmatrix}}_{\mb{S}_{cm_i}}{}^t\bs{\xi}(s_{d_i})
\end{equation} 
Using $m_{d_i}$ and $\bs{\mathcal{I}}_{d_i}$ to denote the mass and inertia tensor of the $i^{th}$ disk expressed in \{U\}, the kinetic energy of all disks is given by:
\begin{equation}
    T_{SD} = \frac{1}{2}\sum_{i=1}^{n_d}{}^u\bs{\xi}\T_{cm_i}\begin{bmatrix}m_{d_i}\mb{I} & \mb{0} \\ 
    \mb{0} & \bs{\mathcal{I}}_{d_i}\end{bmatrix}{}^u\bs{\xi}_{cm_i}
\end{equation}

% \begin{equation}
%     T_{SD} = \frac{1}{2}\sum_{i=1}^{n_d}{}^t\bs{\xi}\T(s_{d_i})\mb{S}\T_{cm_i}\begin{bmatrix}m_{d_i}\mb{I} & \mb{0} \\ 
%     \mb{0} & \bs{\mathcal{I}}_{d_i}\end{bmatrix}\mb{S}_{cm_i}{}^t\bs{\xi}(s_{d_i})
% \end{equation}
%
Using ${}^t\bs{\xi}(s)=\mb{J}_{\xi c}(s)\dot{\mb{c}}$ and \eqref{eq:disk_twist_in_u}, we write $T_{SD}$ as a function of the modal coefficient rates: 
\begin{equation}\label{eq:T_SD}
    T_{SD} = \frac{1}{2}\dot{\mb{c}}\T\,\mb{M}_{SD}\,\dot{\mb{c}}
\end{equation}
where $\mb{M}_{SD}$ is given by:
\begin{equation}
    \mb{M}_{SD} =\sum_{i=1}^{n_d}\mb{J}_{\xi c}\T(s_{d_i})\mb{S}\T_{cm_i}
    \begin{bmatrix}
    m_{d_i}\mb{I} & \mb{0} \\ 
    \mb{0} & \bs{\mathcal{I}}_{d_i}
    \end{bmatrix}
    \mb{S}_{cm_i}\mb{J}_{\xi c}(s_{d_i})
\end{equation}

\subsubsection{Actuation Capstans and Motors}\label{sec:T_A}
%--------------------
\begin{figure}[H]
  \centering
  \includegraphics[width=0.65\columnwidth]{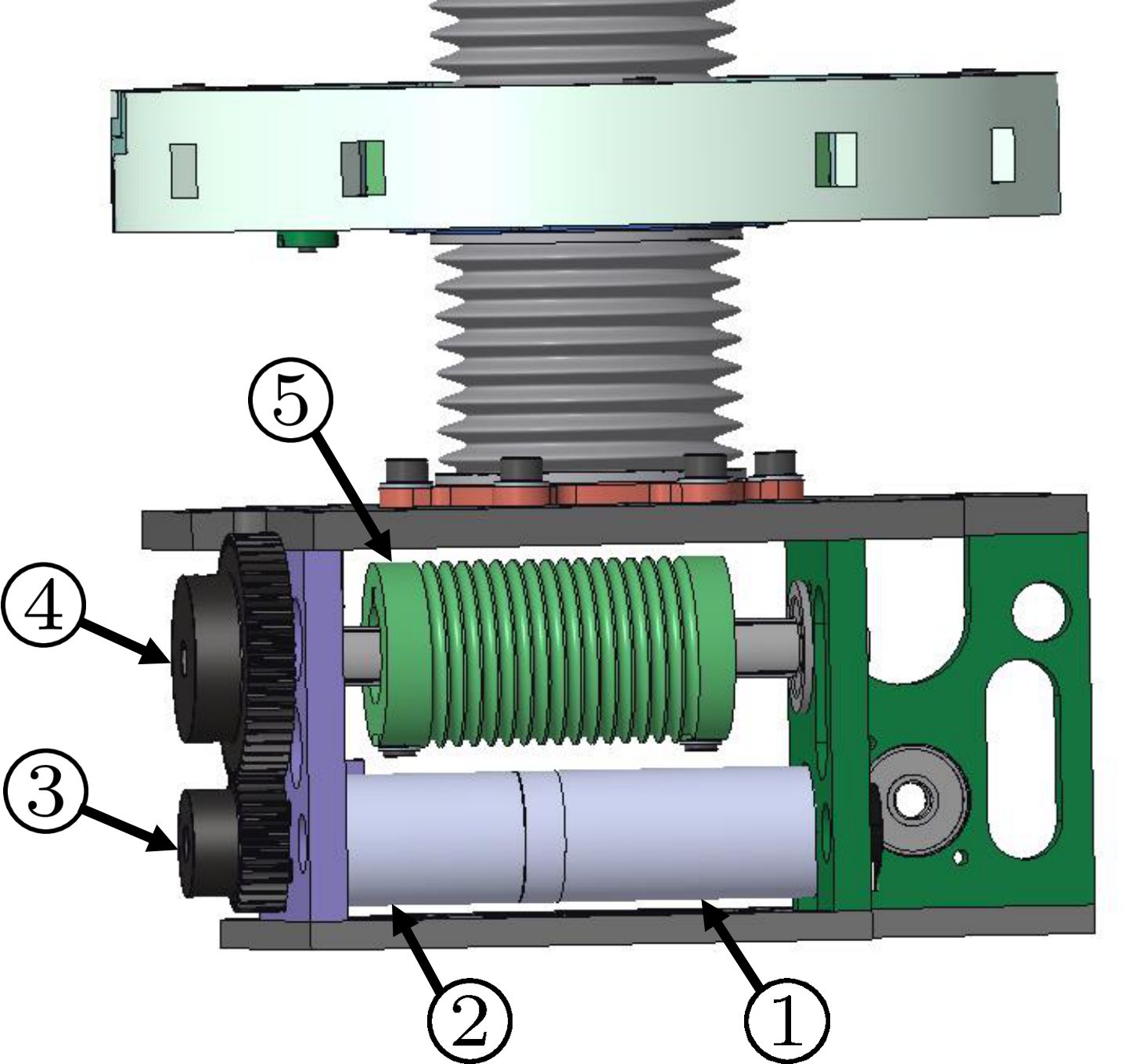}
  \caption{Continuum robot actuation unit: \protect\circled{1} Motor with rotor inertia $J_{m}$, \protect\circled{2} gearhead with inertia $J_{gh}$ and reduction ratio $R_{gh}$, \protect\circled{3} pinion with inertia $J_p$ about the center of rotation, \protect\circled{4} gear with inertia $J_g$ about the center of rotation, \protect\circled{5} shaft and capstan with combined inertia $J_c$. The gear and pinion have a reduction ratio of $R_{gp}$.}\label{fig:actuation_unit}
\end{figure}
%--------------------
Each axis of the continuum robot actuation unit consists of a motor (Fig. \ref{fig:actuation_unit}\circled{1}) with rotor inertia $J_m$ attached to gearhead (Fig. \ref{fig:actuation_unit}\circled{2}) with inertia $J_{gh}$ as viewed by the gearhead input and reduction ratio $R_{gh}$. The gearhead is attached to a pinion (Fig. \ref{fig:actuation_unit}\circled{3}) with inertia $J_p$ which meshes with a gear (Fig. \ref{fig:actuation_unit}\circled{4}) with inertia $J_g$. The gear is rigidly attached to a shaft and capstan (Fig. \ref{fig:actuation_unit}\circled{5}) with combined inertia $J_c$. Using this information, the inertia of the entire actuation chain as viewed by the capstan can be written as:    
\begin{equation}
    J_A = R_{g}^2\left[R_{gh}^2(J_{gh}+J_m) + J_{p}\right] + J_g + J_c
\end{equation}
The kinetic energy of the capstans $T_A$ can be found using $J_A$ and the angular velocity of the capstans $\dot{\mb{q}}\in\realfield{2}$: 
\begin{equation}
    T_A = \frac{1}{2}\dot{\mb{q}}\T\begin{bmatrix}J_A & 0 \\ 0 & J_A\end{bmatrix}\dot{\mb{q}}
\end{equation}
$T_A$ can be written in terms of the modal coefficient rates using:
\begin{equation}\label{eq:T_A}
    T_A = \frac{1}{2}\dot{\mb{c}}\T\mb{M}_A\dot{\mb{c}}
\end{equation}
where $\mb{M}_A$ is given by:
%
% \begin{equation}
%     T_A = \frac{1}{2}\left(\mb{J}_{qc}\dot{\mb{c}}\right)\T\begin{bmatrix}J_A & 0 \\ 0 & J_A\end{bmatrix}\mb{J}_{qc}\dot{\mb{c}}
% \end{equation}
%
\begin{equation}
    \mb{M}_A= \mb{J}_{qc}\T\begin{bmatrix}J_A & 0 \\ 0 & J_A\end{bmatrix}\mb{J}_{qc}
\end{equation}

Using \eqref{eq:T_CB}, \eqref{eq:T_SD}, and \eqref{eq:T_A}, the total kinetic energy of the continuum robot can be written as:
\begin{equation}\label{eq:M}
    T = \frac{1}{2}\dot{\mb{c}}\T\underbrace{\left(\mb{M}_{CB} + \mb{M}_{SD} + \mb{M}_{A}\right)}_{\mb{M}}\dot{\mb{c}}
\end{equation}
Note that, in reality, the capstans also slowly translate along their supporting shafts as they rotate. The kinetic energy from this effect is ignored in this model. 
\subsection{Potential Energy}\label{sec:V}
The total potential energy of the robot can be modeled as the sum of the bending energy $V_B$ of the central backbone, the gravitational potential energy of the central backbone $V_{CB}$, and the gravitational potential energy of the spacer disks $V_{SD}$.
\begin{equation}
    V = V_B + V_{CB} + V_{SD} 
\end{equation}
\corrlab{R1-3.3}{The derivations given in this section are extended to general curvature from the constant curvature models presented in \cite{Roy2016_dynamic_effect_collision}.}
\subsubsection{Elastic Potential Energy}
The bending stiffness matrix of the central backbone can be written as: 
\begin{equation}
    \mb{K} = \operatorname{diag}([EI_x,~EI_y,~JG])
\end{equation}
Where $EI_x$ is the flexural rigidity of the central backbone in the $x$ direction, $EI_y$ is the flexural rigidity in the $y$ direction, and $JG$ is the torsional rigidity of the backbone. Using the stiffness matrix, the potential energy due to bending of the central backbone can be written as: 
\begin{equation}
    V_B = \frac{1}{2}\int_0^L\mb{u}\T(s)\,\mb{K}\,\mb{u}(s)ds
\end{equation}
Using \eqref{eq:curvature}, $V_B$ can be written in terms of $\mb{c}$ as:
\begin{equation}
    V_B = \frac{1}{2}\mb{c}\T\underbrace{\left(\int_0^L\bs{\Phi}\T(s)\mb{K}\bs{\Phi}(s)ds\right)}_{\mb{K}_c}\mb{c}
\end{equation}
% %
% \begin{equation}
%     V_B = \frac{1}{2}\mb{c}\T{\mb{K}_c}\mb{c}
% \end{equation}
% %
\subsubsection{Gravitational Potential Energy}
The gravitational potential energy of the central backbone can be written as:
\begin{equation}
    V_{CB} = -\rho\mb{g}\T\int_0^L {}^0\mb{p}(s)ds
\end{equation}
Where $\mb{g}=[0,0,-9.81]\T$ m/s$^2$ is the acceleration of gravity, ${}^0\mb{p}(s)$ is the position vector of the backbone at arc length $s$, and $\rho$ is the mass per unit length of the backbone. 
\par The gravitational potential energy of the spacer disks can be written as:
\begin{equation}
    V_{SD} = -\sum_{i=1}^{n_d} m_{d_i}\mb{g}\T\left[{}^0\mb{p}(s_{d_i}) + \Rot{0}{t}(s_{d_i}){}^t\mb{p}_{cm_i}\right]
\end{equation}
Where $m_{d_i}$ is the mass of the $i^\text{th}$ disk, ${}^0\mb{p}(s_{d_i})$ is the position of the central backbone at spacer disk anchoring arc length $s_{d_i}$ and $\Rot{0}{t}(s_{d_i})$ was defined in \eqref{eq:local_frame}.
\subsection{Approximate Joint Friction}\label{sec:friction}
%--------------------
\begin{figure}
 \centering
  \includegraphics[width=0.65\columnwidth]{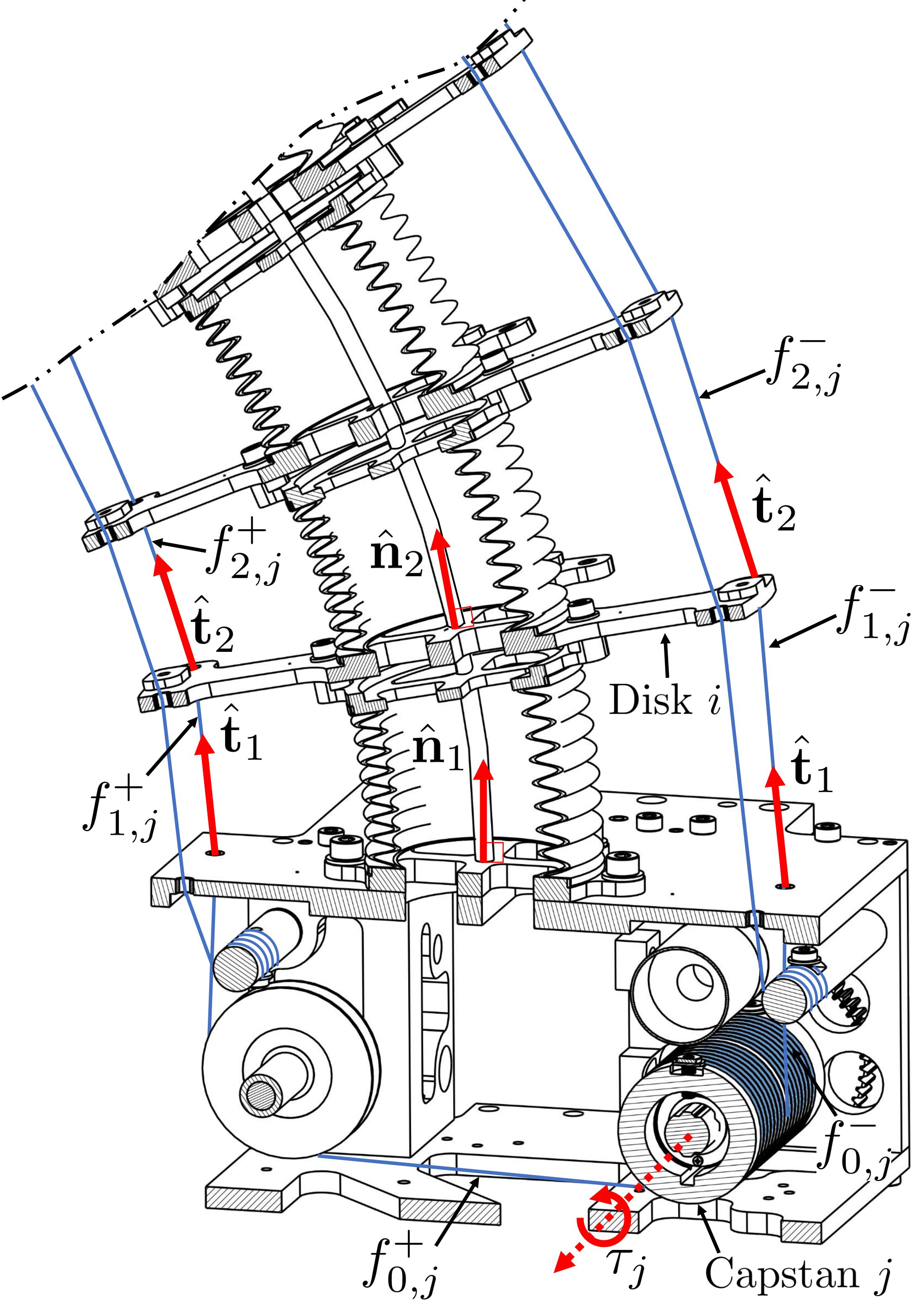}
  \caption{Continuum segment cross section showing the terms used to calculate the joint friction}\label{fig:friction}
\end{figure}
%--------------------
In this section, we will approximate the effect of friction between the actuation tendons and the brass bushings that guide the actuation tendons through the spacer disks. Our approach, which is unique to this paper, uses a simplistic Coulomb friction model that ignores nonlinear, velocity-dependent effects such as Stribeck and viscous friction. As shown in Figs. \ref{fig:shape_sensing} and \ref{fig:friction}, the robot's actuation tendons are wrapped around a motorized capstan and then routed through holes in the spacer disks. Once the tendon reaches the end-disk, they are rerouted back through the spacer disks using a pulley (see Fig. \ref{fig:shape_sensing}) and terminated on a fixed shaft. This is repeated on the same capstan, but separated by 180$^\circ$ to control bending in both directions in the same plane. To ensure the tendons are always in tension, each side of the tendon are pre-tensioned to $f_{pl} = 208$ N. 
\par The pretensions cannot create a moment about the capstan because they act in equal and opposite directions on the capstan\footnote{In practice, any small deviations in the pretension are not enough to backdrive the capstans.}. The pretension does, however, affect the friction between the tendons and the bushings because it contributes to the normal force. When torque $\tau_j$ is applied to capstan $j\in [1,2]$, the tension in the tendon being pulled by the capstan is $f^+_{0,j}$ as shown in Fig. \ref{fig:friction}. On the release side of the capstan $j$, the tendon tension is reduced to $f^-_{0,j}$. The values of $f^+_{0,j}$ and $f^-_{0,j}$ can be found from $\tau_j$ and $f_{pl}$ using:  
\begin{equation}
    f^+_{0,j} = \frac{|\tau_j|}{r_c} + f_{pl},\quad f^-_{0,j} = \max\left(f_{pl}-\frac{|\tau_j|}{r_c},~0\right)
\end{equation}
As the tendon passes through the bushing in disk $i$, the tendon tension is reduced by the friction force between the bushing and the tendon ($\eff^+_{i,j}$ or $\eff^-_{i,j}$): 
\begin{equation}\label{eq:friction}
    f^+_{i,j} = f^+_{i-1,j} - \eff^+_{i,j},\quad f^-_{i,j} = f^-_{i-1,j} - \eff^-_{i,j}
\end{equation}
In this section, we will use the convention that $i = 1$ at the segment base as shown in Fig. \ref{fig:friction}. The friction forces depend on the magnitude of normal force between the tendon and the bushing ($\mathfrak{n}^+_{i,j}$ or $\mathfrak{n}^-_{i,j}$) which can be found by projecting the tendon forces onto the plane of the disk:  
\begin{equation}
\begin{split}
    \mathfrak{n}^+_{i,j} =& \|\mb{P}_i(f^+_{i-1,j}\uvec{t}_{i-1} + f^+_{i,j}\uvec{t}_i)\|,\\
    \quad \mathfrak{n}^-_{i,j} =& \|\mb{P}_i(f^-_{i-1,j}\uvec{t}_{i-1} + f^-_{i,j}\uvec{t}_i)\| 
\end{split}
\end{equation}
where $\mb{P}_i$ is the matrix that projects vectors onto the plane of the disk ($\mb{P}_i = \mb{I} - \uvec{n}_i\uvec{n}\T_i$) and  $\uvec{n}_i$ is a unit vector normal to the disk ($3^{rd}$ column of $\Rot{0}{t}(s_{d,i})$). Also, $\uvec{t}_i$ and $\uvec{t}_{i-1}$ are unit vectors in the direction of the tendon segments after disks $i$ and $i-1$, respectively. These unit vectors can be found from the geometry of the disks and the local disk frames.  
\par Given the tension $f^+_{i-1,j}$ or $f^-_{i-1,j}$, the tension $f^+_{i}$ or $f^-_{i}$ can be calculated using a constrained minimization approach. The following optimization problem finds the minimum value the tension subject to the constraints that the wire stays in tension and the friction force is proportional to the normal force by the friction coefficient $\mu$: 
\begin{equation}\label{eq:tension_optim}
\begin{aligned}
     \min~&~f_{i,j}^2 \\
     s.t.~&~ 0 \leq f_{i,j}, \quad \eff_{i,j} = \mu \mathfrak{n}_{i,j}
\end{aligned}
\end{equation}
This optimization problem must be solved starting from $i=1$ (i.e., the segment base) and progressing at each section of tendon until the end-disk and back from the end-disk to the termination point. In our approach, the constrained minimization problem given in \eqref{eq:tension_optim} is solved using the sequential quadratic programming algorithm implemented in MATLAB's \textit{fmincon} function. At each disk, after $f^+_i$ and $f^-_i$ is known, the friction force can be found using \eqref{eq:friction}. 
\par Once the friction force at each bushing is known, the total friction force experienced by capstan $j$ is the sum of each value of $\eff^+_{i,j}$ and $\eff^-_{i,j}$:  
\begin{equation}
    \eff_{j} = \sum_{i=1}^{2(n_d + 1)}\eff^+_{i,j} + \sum_{i=1}^{2(n_d + 1)}\eff^-_{i,j}
\end{equation}
The torque on the capstans due to friction can now be found using the capstan radius $r_c$:
\begin{equation}
    \bs{\tau}_{F} = \operatorname{sign}(\dot{\mb{q}})r_c\begin{bmatrix}\eff_{1} & \eff_{2}\end{bmatrix}\T
\end{equation}
For the purposes of simulation, the $\operatorname{sign}(\dot{\mb{q}})$ function was replaced with $\tanh(10\dot{\mb{q}})$ to not introduce discontinuities into the dynamic model which can cause issues with numerical differential equation solvers. Using $\mb{J}_{qc}$, the generalized forces due to friction can be found from $\bs{\tau}_{F}$ using:
\begin{equation}
    \bs{\kappa}_{fric} = \mb{J}_{qc}\T\bs{\tau}_{F}
\end{equation}
\subsection{Equations of Motion}
Using the total kinetic energy of the segment from section \ref{sec:T} and the total potential energy of the segment from section \ref{sec:V}, the Lagrangian $\mathcal{L}$ can be written as:
\begin{equation}
    \mathcal{L} = T-V
\end{equation}
Using $\mathcal{L}$, the segment's equations of motion can be found using the Euler-Lagrange equation:
\begin{equation}\label{eq:lagrange_first_def}
    \frac{d}{dt}\frac{\partial \mathcal{L}}{\partial \dot{\mb{c}}} - \frac{\partial \mathcal{L}}{\partial \mb{c}} = \bs{\kappa}
\end{equation}
Where $\bs{\kappa}$ are the non-conservative generalized forces:
\begin{equation}\label{eq:non_cons_gen_force}
    \bs{\kappa} = \mb{J}_{qc}\T\bs{\tau} + \mb{J}_{\xi c}\T(s_c)\,\mb{w}_c - \bs{\kappa}_{fric}
\end{equation}
In this equation $\bs{\tau}\in\realfield{2}$ are capstan moments and $\mb{w}_c\in\realfield{6}$ is the external wrench applied to the robot at arc length $s_c$. In this paper, we use the convention that a wrench consists of a force $\mb{f}\in\realfield{3}$ proceeding a moment $\mb{m}\in\realfield{3}$ (i.e., $\mb{w}_c = [\mb{f}_c\T,~\mb{m}_c\T]\T$). Expanding \eqref{eq:lagrange_first_def} results in\footnote{Derivation details are provided in the appendix}:  
\begin{equation}\label{eq:EOM}
    \mb{M}\ddot{\mb{c}}+\mb{N}\dot{\mb{c}} + \frac{\partial V}{\partial \mb{c}} = \bs{\kappa}
\end{equation}
Where $\mb{M}$ is the generalized inertia matrix given in \eqref{eq:M} and $\mb{N}$ is the centrifugal/Coriolis matrix. The derivation of $\mb{N}$ and $\frac{\partial V}{\partial \mb{c}}$ is given in the appendix.

\section{Modal-Space Momentum\\ Observer Collision Detection}
\subsection{Observer Formulation}
In this section, we adapt the momentum observer framework presented in \cite{DeLuca2003_GMO} to our modal approach and present our implementation of the observer. The momentum observer approach relies on a detection of a change in the robot's generalized momentum as a means to detect an incident of collision. The generalized momentum $\mb{p}$ is defined as:
\begin{equation}\label{eq:generalized_momentum}
    \mb{p} = \frac{\partial \mathcal{L}}{\partial \dot{\mb{c}}} = \mb{M}\dot{\mb{c}}
\end{equation}
Since the contact impulse is related to a temporal change in momentum, we calculate $\dot{\mb{p}}$:
\begin{equation}
    \dot{\mb{p}} = \dot{\mb{M}}\dot{\mb{c}} + \mb{M}\ddot{\mb{c}}
\end{equation}
Substituting the solution for $\mb{M}\ddot{\mb{c}}$ from \eqref{eq:EOM} results in:
\begin{equation}\label{eq:p_dot1}
    \dot{\mb{p}} = \dot{\mb{M}}\dot{\mb{c}} - \mb{N}\dot{\mb{c}} - \frac{\partial V}{\partial \mb{c}} + \bs{\kappa}
\end{equation}
Since the matrix $\dot{\mb{M}} - 2\mb{N}$ is skew-symmetric as shown in \cite{Koditschek1984_skew_symmetry}, we use \mbox{$\dot{\mb{M}} - 2\mb{N} = -\left(\dot{\mb{M}}-2\mb{N}\right)\T$} and solve for $\dot{\mb{M}}$. This results in:
\begin{equation}
    \dot{\mb{M}} = \mb{N}\T + \mb{N} 
\end{equation}
Using this result along with \eqref{eq:non_cons_gen_force}, equation \eqref{eq:p_dot1} can now be expressed as: 
% %
% \begin{equation}
%     \dot{\mb{p}} = \left(\mb{N}\T + \mb{N}\right)\dot{\mb{c}} - \mb{N}\dot{\mb{c}} - \frac{\partial V}{\partial \mb{c}} + \bs{\kappa}
% \end{equation}
% %
\begin{equation}
    \dot{\mb{p}} = \mb{N}\T\dot{\mb{c}} - \frac{\partial V}{\partial \mb{c}} + \mb{J}_{qc}\T\bs{\tau} -\bs{\kappa}_{fric}+\mb{J}_{\xi c}\T(s_c)\,\mb{w}_c
\end{equation}
Since we do not know the contact wrench $\mb{w}_c$, we define a residual $\mb{r}$ representing the effect of an unknown contact impulse on the system. Since we plan to estimate $\mb{r}$, the momentum temporal rate is rewritten as:
\begin{equation}\label{eq:p_dot_estimate1}
    \dot{\widetilde{\mb{p}}} = \mb{N}\T\dot{\mb{c}} - \frac{\partial V}{\partial \mb{c}} + \mb{J}_{qc}\T\bs{\tau} -\bs{\kappa}_{fric}+\mb{r}
\end{equation}
In the following derivation, we assume that a contact estimation algorithm is updated at every sensory acquisition cycle $i$, but the value of $\mb{r}$ is updated from the preceding cycle:
\begin{equation}\label{eq:p_dot_estimate2}
    \dot{\widetilde{\mb{p}}}_i = \left(\mb{N}\T\dot{\mb{c}} - \frac{\partial V}{\partial \mb{c}} + \mb{J}_{qc}\T\bs{\tau} -\bs{\kappa}_{fric}\right)_i+\mb{r}_{i-1}
\end{equation}
To reject sensory noise, we assume that $\mb{r}_{i-1}$ includes estimate updates from the previous $n$ sensory cycles. At any $k^{th}$ cycle ($k\in[1,n_w]$, $n_w>>n$) we define the momentum residual as:
\begin{equation}\label{eq:GMO1}
    \mb{r}_k = \mb{K}_o\left[\mb{p}_k- \mb{p}_1-\sum_{j=k-n}^{k}\dot{\widetilde{\mb{p}}}_j\,\text{d}t\right], \quad k\in[1,n_w] 
\end{equation}
In the above definition, $dt$ is the sensory update cycle time, $n_w$ denotes the total number of cycles where the estimator is allowed to run before it is reset. This is needed to avoid false positives due to cumulative summation/integration effects of measurement noise and model error. The matrix $\mb{K}_o$ is a diagonal positive definite  gain matrix determining the dynamics of the momentum residual as discussed in section \ref{sec:estimator_dynamics}, $\mb{p}_1$ is the initial generalized momentum at the beginning of a contact estimation span $k=1$.
\par Substituting \eqref{eq:p_dot_estimate2} in \eqref{eq:GMO1} results in the explicit form of the update law for the residual:
\begin{equation}\label{eq:GMO2}
\begin{split}
    \mb{r}_k = &\mb{K}_o\Biggl[\mb{M}_k\dot{\mb{c}}_k- \mb{p}_1-\\
    \sum_{j=k-n}^{k}\Biggl(
    &\left(\mb{N}\T\dot{\mb{c}} - \frac{\partial V}{\partial \mb{c}} + \mb{J}_{qc}\T\bs{\tau} - \bs{\kappa}_{fric}\right)_j+\mb{r}_{j-1}\Biggr)\,\text{d}t\Biggr]
\end{split}
\end{equation}
%===================================================================
\subsection{Estimator Dynamics \& Stability}\label{sec:estimator_dynamics}
In this section, we will repeat a result first introduced by De Luca and Mattone \cite{DeLuca2003_GMO} which proves that the GMO is a stable, first-order, filtered estimate of the generalized forces due to contact. To investigate the estimation dynamics, we will cast \eqref{eq:GMO1} into the continuous time domain and then take the derivative with respect to time:
\begin{equation}
    \dot{\mb{r}} = \mb{K}_o\left[\dot{\mb{p}}(t)-\frac{d}{dt}\int_0^t\dot{\widetilde{\mb{p}}}(\sigma)d\sigma\right]
\end{equation}
This equation simplifies to:
\begin{equation}
    \dot{\mb{r}} = \mb{K}_o\left[\dot{\mb{p}}(t)-\dot{\widetilde{\mb{p}}}(t)\right]
\end{equation}
Plugging \eqref{eq:p_dot1} and \eqref{eq:p_dot_estimate2} in for $\dot{\mb{p}}$ and $\dot{\widetilde{\mb{p}}}$, respectively, and simplifying yields:
\begin{equation}\label{eq:r_dot_time}
    \dot{\mb{r}}(t) = \mb{K}_o\left[\bs{\kappa}_c(t) - \mb{r}(t)\right]
\end{equation}
where $\bs{\kappa}_c(t)=\mb{J}\T_{\xi c}(s_c)\mb{w}_c$. In the Laplace domain, the $i^\text{th}$ element of \eqref{eq:r_dot_time} can be written as:
\begin{equation}
    sr_i(s) = K_{o,i}\left[\kappa_{c,i}(s) - r_i(s)\right]
\end{equation}
Solving for $\kappa_{c,i}$ yields:
\begin{equation}
    r_i(s) = \frac{K_{o,i}}{s+K_{o,i}}\kappa_{c,i}(s)
\end{equation}
This equation can be re-written to reveal the time constant of the filter:
\begin{equation}\label{eq:transfer_func}
   r_i(s) = \frac{1}{\frac{1}{K_{o,i}}s+1}\kappa_{c,i}(s)
\end{equation}
As shown in \cite{DeLuca2003_GMO}, for a positive estimation gain $K_{o,i} > 0$, the transfer function $ r_i(s)/\kappa_{c,i}(s)$ given in \eqref{eq:transfer_func} has one stable (i.e. negative) pole at $s = -K_{o,i}$. \remind{R1-7.1}{Therefore, in ideal scenarios, the estimation signal $r_i$ will approach the true generalized force $\kappa_{c,i}$ with time constant  $1/K_{o,i}$. From this analysis, it can be seen that selecting a gain is a balancing act between responsiveness of the estimator and noise filtering. Selecting a higher $K_{o,i}$ value will create a more responsive estimator, but the estimator will be more sensitive to noise. Conversely, a lower gain will more effectively filter noise, but will have a delay in the estimation.}
%------------------------------------------------------------

\section{Wrench estimation}
 The output of the momentum observer $\mb{r}$ is an estimate of the generalized forces due to contact. To determine the corresponding estimate of the applied wrench, $\widetilde{\mb{w}}_c$, the following equation must be solved for $\widetilde{\mb{w}}_c$:
\begin{equation}\label{eq:w_e_estimate}
    \mb{r} = \mb{J}_{\xi c}\T(s_c)\widetilde{\mb{w}}_c
\end{equation}
In scenarios where the location of contact is known (e.g., from a sensor \cite{Abah2019_sensor_disk,Abah2022_sensor_disk}) and $\mb{J}_{\xi c}\T(s_c)\in\realfield{6\times6}$ is full rank, $\widetilde{\mb{w}}_c$ can be found by inverting $\mb{J}_{\xi c}\T(s_c)$. 
\subsection{Constrained minimization-based wrench estimation}
However, the continuum segment shown in Fig. \ref{fig:shape_sensing} cannot instantaneously translate along the local $\uvec{z}$ axis due to its nickel-titanium backbone. Therefore, for this robot, $\operatorname{rank}\left[\mb{J}_{\xi c}(s_c)\right] \leq 5$. This has two implications 1) \eqref{eq:w_e_estimate} can no longer be solved by simple inversion of $\mb{J}_{\xi c}\T(s_c)$ and 2) if $\mb{w}_c \in \operatorname{Null}\left[\mb{J}_{\xi c}\T(s_c)\right]$ the wrench is fundamentally insensible. Therefore, we introduce a constrained minimization based approach to the wrench estimation problem: 
\begin{equation}\label{eq:weighed_lsq}
\begin{aligned}
 \min_{\widetilde{\mb{w}}_c}\quad & \widetilde{\mb{w}}_c\T\mb{W}\widetilde{\mb{w}}_c\\
\textrm{s.t.}\quad &~\mb{r} = \mb{J}_{\xi c}\T(s_c)\widetilde{\mb{w}}_c 
\end{aligned}
\end{equation}
In this equation, $\mb{W}\in\realfield{6\times6}$ is a diagonal weight matrix. This minimization problem will find the weighted minimum-norm solution to \eqref{eq:w_e_estimate}.
\subsection{Incorporating Knowledge of the Contact Type}
In some scenarios, information about the type of external contact is known which imposes constraints on the components of the applied wrench. For instance, if the contact is assumed to be a point contact, it cannot impart a moment to the segment and the moment components of $\widetilde{\mb{w}}_c$ can be assumed to be zero. For a comprehensive list of contact constraints, see \cite{Mason1985_robot_hands}. The approach presented here is a modification of the constrained optimization approach initially presented in \cite{Xu2008_continuum_collision}. Additionally, due to the assumption of infinite torsional rigidity and inextensibility of the central backbone, any applied moments/forces about/along the local $z$ axis cannot alter the generalized momentum of the robot and will therefore be insensible by the momentum observer and can be assumed to be zero. These assumptions can be incorporated in the wrench estimation using the constraint matrix $\mb{A}\in\realfield{n_c\times6}$ where $n_c$ is the number of components of $\widetilde{\mb{w}}_c$ that can be assumed to be zero:     
\begin{equation}\label{eq:wrench_est}
\begin{aligned}
 \min_{\widetilde{\mb{w}}_c}\quad & \widetilde{\mb{w}}_c\T\mb{W}\widetilde{\mb{w}}_c\\
\textrm{s.t.} \quad & \mb{r} = \mb{J}_{\xi c}\T(s_c)\widetilde{\mb{w}}_c, \quad  \mb{A}\widetilde{\mb{w}}_c = \mb{0}
\end{aligned}
\end{equation}
For the simulations and experiments in this paper, the applied wrench is assumed to be a point contact and the force along the local $z$ axis is assumed to be zero for the reasons described above. Therefore, we will use the following $\mb{A}$ matrix:
\begin{equation}\label{eq:A}
    \mb{A} = %\begin{bmatrix}
            % 0 & 0 & 1 & 0 & 0 & 0 \\
            % 0 & 0 & 0 & 1 & 0 & 0 \\
            % 0 & 0 & 0 & 0 & 1 & 0 \\
            % 0 & 0 & 0 & 0 & 0 & 1 \\
            %  \end{bmatrix}
            [\mb{0}_{4\times 2},\mb{I}_4]
\end{equation}
\corrlab{R2-5}{For the simulations and experiments in this paper, \eqref{eq:wrench_est} was implemented in MATLAB 2023a and was compiled using MATLAB coder. During the experiments shown in Fig. \ref{fig:exp_results}(a), \eqref{eq:wrench_est} was solved with a median execution time of 0.0390 seconds and a maximum execution time of 0.1297 seconds. The computer used for this analysis has 32 GB of RAM and an Intel\texttrademark~i7-13700 processor. The operating system was Windows 10.}
\section{Effect of State \\Uncertainty on Wrench Estimation}\label{sec:uncertainty}
In the following, we consider the effect of parameter and model uncertainty on the GMO method. Understanding how uncertainty in the segment's dynamic state affects the output of GMO is critical to characterizing the expected performance of the method. We therefore investigate how an unmodeled error in the state vector will affect the GMO residual. 
\par Since our formulation is in modal space, we define the segment state vector as:
\begin{equation}
    \mb{x} = [\mb{c}\T,~\dot{\mb{c}}\T]\T \in \realfield{12}
\end{equation}
The effect of an unmodeled change $\Delta\mb{x}$ on the $i^\text{th}$ element of $\mb{r}$ can be expressed using a Taylor series expansion: 
\begin{multline}
    r_i(\mb{x}+\Delta\mb{x}) = r_i(\mb{x}) + \sum_{j=1}^{12}\frac{\partial r_i(\mb{x})}{\partial x_j}\Delta x_j \\ + \frac{1}{2}\sum_{k=1}^{12}\sum_{j=1}^{12}\frac{\partial^2 r_i(\mb{x})}{\partial x_k\partial x_j}\Delta x_k\Delta x_j+\dots
\end{multline}
In this work, we will approximate the effect of an unmodeled deviation in dynamic state $\Delta\mb{x}$ using a first-order approximation of the Taylor series:   
\begin{equation}\label{eq:r_x_deltax}
    \mb{r}(\mb{x}+\Delta\mb{x}) \approx \mb{r}(\mb{x}) + \frac{d\mb{r}(\mb{x})}{d\mb{x}}\Delta\mb{x}
\end{equation}
The derivative of $\mb{r}$ with respect to $\mb{x}$ can be found by differentiating \eqref{eq:GMO1}:
\begin{equation}\label{eq:dr_dx}
    \frac{d\mb{r}_k(\mb{x})}{d\mb{x}} = \mb{K}_o\left[ \frac{d\mb{p}_k}{d\mb{x}} - \sum_{j=k-n}^k\frac{d\dot{\widetilde{\mb{p}}}_j}{d\mb{x}}\text{d}t\right]    
\end{equation}
First, we will calculate $\frac{d\mb{p}_k}{d\mb{x}}$. The $i^\text{th}$ element of $\mb{p}_k$ can be represented using the sum:
\begin{equation}
    p_i = \sum_{l=1}^6M_{il}\dot{c}_l
\end{equation}
Note that in this equation and the subsequent analysis, the subscript $k$, indicating the sample number, has been dropped for clarity. Using this summation representation of $\mb{p}$, the $ij^\text{th}$ element of $\frac{d\mb{p}}{d\mb{x}}$ can be written as:
\begin{equation}
    \frac{\partial p_i}{\partial x_j} = \sum_{l=1}^6\frac{\partial M_{il}}{\partial x_j}\dot{c}_l + M_{il}\frac{\partial \dot{c}_k }{\partial x_j}
\end{equation}
In our implementation, the partial derivatives of the elements of $\mb{M}$ are computed using a central finite differences approximation. The partial derivative of $\dot{c}_l$ with respect to $x_j$ can be computed simply as:
\begin{equation}
    \frac{\partial \dot{c}_l }{\partial x_j} = \begin{cases} 
                                          0, & \dot{c}_l \neq x_j \\
                                          1, & \dot{c}_l = x_j
                                          \end{cases}
\end{equation}
\par Next, we will compute the derivative of $\dot{\widetilde{\mb{p}}}_j$  with respect to the state vector by differentiating \eqref{eq:p_dot_estimate2} with respect to $\mb{x}$:
\begin{equation}
\begin{split}
    \frac{d\dot{\widetilde{\mb{p}}}_j}{d\mb{x}} = \frac{d}{d\mb{x}}\left(\mb{N}_j\T\dot{\mb{c}}_j\right) - \frac{d}{d\mb{x}}\left[\frac{\partial V}{\partial \mb{c}}\right]_j + \cancelto{0}{\frac{d}{d\mb{x}}\mb{J}_{qc}\T\bs{\tau}}_j \\
    -\frac{d}{d\mb{x}}\bs{\kappa}_{fric,j} + \frac{d}{d\mb{x}}\mb{r}_{j-1}
\end{split}
\end{equation}
Because $\mb{J}_{qc}$ is not a function of $\mb{c}$ or $\dot{\mb{c}}$, the generalized forces due to the capstan torques cancel to zero. The term $\frac{d}{d\mb{x}}\mb{r}_{j-1}$ is the value of $\frac{d}{d\mb{x}}\mb{r}$ from the previous time step. In our implementation, the other derivatives in this equation are computed via central finite differences.  
\par Given a characterized uncertainty in $\mb{x}$, \eqref{eq:r_x_deltax} can be used to predict the associated uncertainty in the GMO estimation residual $\mb{r}$ and subsequently the estimated applied wrench. This can be used to set thresholds on the elements of $\mb{r}$ where, if the value is less than the threshold, it cannot be determined if the residual is due to uncertainty in the dynamics or due to a small contact force and therefore should be ignored in practical applications. Conversely, if the value is above the threshold, the residual cannot be fully explained by dynamic model errors and thus the robot should be considered in a contact state. To establish these thresholds, we plug $\mb{x} = \mb{0}$ into \eqref{eq:r_x_deltax} and set $\Delta\mb{x}$ to the characterized uncertainty in the dynamic state. Because \eqref{eq:r_x_deltax} is a first-order truncation of the Taylor series, it is inherently an approximation and there may be small differences between the predicted and actual observer residuals for a given error in $\mb{x}$. 

% \begin{equation}
%     \left[\frac{d}{d\mb{x}}\mb{N}\T\dot{\mb{c}}\right]_{ij} = \sum_{k=1}^6\frac{\partial N\T_{ik}}{\partial x_j}\dot{c}_k + N\T_{ik}\frac{\partial \dot{c}_k }{\partial x_j}
% \end{equation}

% %
% \begin{equation}
%     \frac{d}{d\mb{x}}\frac{\partial V}{\partial \mb{c}} =
%     \begin{bmatrix}
%      \frac{\partial^2 V}{\partial c_i \partial c_j} & \mb{0} \\ \mb{0} & \mb{0}                                                               
%     \end{bmatrix}
% \end{equation}

\section{Joint Force/Torque Deviation}
One intuitive method for contact detection/estimation is the joint force/torque deviation (JFD) method. This method was proposed in some of the earliest works on robot contact estimation (e.g., \cite{Suita1995_JFD}) and has been applied to continuum robots for contact detection for static scenarios in \cite{Bajo2010_continuum_collision}. This method is also referred to in the literature as the \emph{direct estimation} method (e.g., \cite{haddadin2017_collsion_review}). The key idea of this method is to rearrange the robot's full dynamic model \eqref{eq:EOM} to directly solve for the generalized forces due to contact:
\begin{equation}
    \widetilde{\bs{\kappa}}_c = \mb{M}\ddot{\mb{c}}+\mb{N}\dot{\mb{c}} + \frac{\partial V}{\partial \mb{c}} - \mb{J}\T_{qc}\mb{\tau} + \bs{\kappa}_{fric}
\end{equation}
Using $ \widetilde{\bs{\kappa}}_c$, the method described in \eqref{eq:wrench_est} can be used to estimate the applied wrench during the contact. The major drawback of the JFD is the need to compute the second derivative of the modal coefficients. Unlike the noise filtering nature of the GMO, taking the numerical second derivative of $\mb{c}$ is a noise amplification process. Significant care needs to be taken to apply low-pass filters to the numerical derivatives of $\mb{c}$ and to the estimation signal $\widetilde{\bs{\kappa}}_c$. 
\par In the experimental evaluation section of this paper, the JFD method is used as a baseline of comparison for the performance of the GMO.  
\section{Dynamic Model Calibration}\label{sec:calibration}
\begin{table*}[]
\caption{Dynamic Model Parameters}\label{tab:dynamic_params}
\fontsize{9}{12}\selectfont
\begin{tabular}{|c|c|c|}
\hline
$L=300.65 mm$ & $r_c=15.255 mm$ & $\gamma=2.83 mm$ \\ \hline
$\rho=83.1532 g/m$ & $r=2 mm$ & $n_d=6$ \\ \hline
$JG=1$ & $J_A=14.323 gm^2$ & $\mb{g}=[0,~0,~-9.81]\T$ $m/s^2$ \\ \hline
$\mb{s}_d={[}53.08, 102.62, 153.16, 203.70, 254.24, L{]} mm$ & $m_{d,i}=308.81 g,~i \in [1, n_d-1]$ & $m_{d,i}=743.12 g,~i = n_d$ \\ \hline
\multicolumn{3}{|c|}{
    \parbox{70mm}{
        \noindent \hspace{-30mm} 
        \begin{equation*}
            \begin{split}
                \bs{\mathcal{I}}_{d,i}=&
                \begin{bmatrix}0.5211 & 0.0024 & -0.0002 \\ 
                    0.0024 & 0.5273 & 0.0007 \\
                    -0.0002 & 0.0007  & 0.9934
                \end{bmatrix}~gm^2 \\
            &\quad i \in[1, n_d-1] 
            \end{split}
        \end{equation*}
    }, \quad  
    \parbox{70mm}{
        \noindent \hspace{-30mm} 
        \begin{equation*}
            \begin{split}
                \bs{\mathcal{I}}_{d,i}&= 
                \begin{bmatrix} 
                    1.1580 & -0.0357 & 0.0001 \\ 
                    -0.0357 & 1.4871 & 0 \\ 
                    0.0001  & 0 & 2.0564 
                \end{bmatrix}~gm^2 \\
            & i = n_d
        \end{split}
        \end{equation*}
    }  %end of parbox
}  %end of multicolumn
\\ 
\hline
\parbox{50mm}{
\noindent \hspace{-30mm} 
\begin{equation*}
    \begin{split}
        \mb{p}_{cm,i}= &\begin{bmatrix}-0.0739\\~0.1954\\~5.7456\end{bmatrix} mm , \quad i \in [1,n_d-1]\\
    \end{split}
\end{equation*}
}
&
\parbox{50mm}{
\noindent \hspace{-30mm} 
\begin{equation*}
    \begin{split}
        \mb{p}_{cm,i}= &\begin{bmatrix}-0.0133\\~0\\~20.8966\end{bmatrix} mm, \quad i = n_d
    \end{split}
\end{equation*}
}
& 
$f_{pl}=208 N$ \\ \hline
\end{tabular}
\end{table*}

\par The dynamic model presented in section \ref{sec:dynamics} contains numerous parameters (masses, inertias, etc.). Most of these parameters can be found from physical measurements, known from the robot geometry, or estimated from CAD software. These parameters are summarized in Table \ref{tab:dynamic_params}. Note that because of the assumption of zero torsional deflections implicit to the formulation of \eqref{eq:curvature}, the value used for $JG$ does not affect the system dynamics. Some parameters, however, need to be calibrated from experimental data. These parameters were chosen to be the backbone flexural rigidities, $EI_x$ and $EI_y$, and friction coefficients $\mu_1$ and $\mu_2$.
\par To calibrate these parameters, the segment's capstans were commanded to move with a chirp trajectory that starts from $f_1 = 0.01$ Hz and linearly increases to $f_0 = 0.25$ Hz in $t_f = 6$ seconds, has an amplitude of $q_{max} = 360^\circ$ (which corresponds to 42 degrees of segment bending), and a phase offset of $\phi_0 = 90^\circ$. The chirp trajectory is given by:
\begin{equation}
q(t) = q_{max}\sin\left[2\pi\left(\left(\frac{f_1 - f_0}{2t_f}\right)t^2 + f_0t\right) + \phi_0\right]    
\end{equation}
\par During the experiment, the string encoders were read using a Teensy 4.1 microcontroller and sent to the central control computer using user datagram protocol (UDP) communication. For more information on the string encoders see \cite{Orekhov2020_workshop,Orekhov2023_shape_sensing}. The segment's motors (Maxon\texttrademark~DCX22L GB KL) were controlled using Maxon\texttrademark~ESCON 50/5 motor drivers, Sensoray\texttrademark~526 encoder reading cards, and a Diamond Systems\texttrademark~Aries PC/104 CPU running Ubuntu 14.04 with the PREEMPT-RT real-time patch. The segment motors were current controlled and the commanded current was used to estimate the capstan torque using the motor's torque current constant, the rated efficiency of the motor and gearhead (Maxon\texttrademark~GPX22HP 62:1), and the reduction ratios of the gearhead and the additional gearing (Fig. \ref{fig:actuation_unit}-\circled{3} and \circled{4}). The encoders and capstan torque readings are sent to the central control computer using UDP communication. The central control computer runs Ubuntu 18.04 with the Melodic distribution of the Robot Operating System (ROS). For the duration of the experiment, data was recorded from the motor encoders, motor torques, and string encoders using ROS topics at an average sampling rate of $\sim$100 Hz.
\par Using the recorded starting pose of the segment and the recorded capstan torques, the segment's motion was simulated using MATLAB's implementation of the Dormand-Prince version of the Runge-Kutta method (\emph{ode45}) first described in \cite{dormand_prince1980_ode45}. Using the recorded modal coefficients from this experiment, the backbone flexural rigidities and friction coefficients were calibrated using the following constrained minimization problem:  
\begin{equation}
\begin{aligned}
    \bs{\alpha} &=  \operatorname{argmin} \sum_{i = 1}^{n}\frac{1}{2}(\mb{c}_i-\widetilde{\mb{c}}_i)\T(\mb{c}_i-\widetilde{\mb{c}}_i)\\
    s.t.~ & ~ \mb{0} < \bs{\alpha} < \bs{\alpha}_{max} 
\end{aligned}
\end{equation}
\par In this equation $\bs{\alpha} = [EI_x,EI_y,\mu_1,\mu_2]\T$, $\mb{c}_i$ is the experimentally determined modal coefficients at sample $i$, and $\widetilde{\mb{c}}_i$ is the simulated modal coefficients at sample $i$. Since we know the geometric parameters of the robot exactly, the uncertainty in the moments of inertia is very small (negligible). The primary source of uncertainty is due to the Young's modulus of the combined superelastic NiTi and bellow backbone system. We used $\bs{\alpha}_{max} = [2.5,2.5,1,1]\T$ during the calibration process after some experimental tuning. This minimization problem was solved using the interior point algorithm implemented in MATLAB's \emph{fmincon} function. The calibrated parameters are summarized in Table \ref{tab:calib_results}. 
\begin{table}[ht]
    \centering
    \caption{Calibrated dynamic model parameters}
    \begin{tabular}{@{}cccc@{}} 
        \toprule
        $EI_x$ & $EI_y$ & $\mu_1$ & $\mu_2$ \\ \midrule
        1.1440 & 1.0373 & 0.0312  & 0.1637 \\ \bottomrule
    \end{tabular}
    \label{tab:calib_results}
\end{table} 
\par
\begin{figure}[htbp]
  \centering
  \includegraphics[width=0.99\columnwidth]{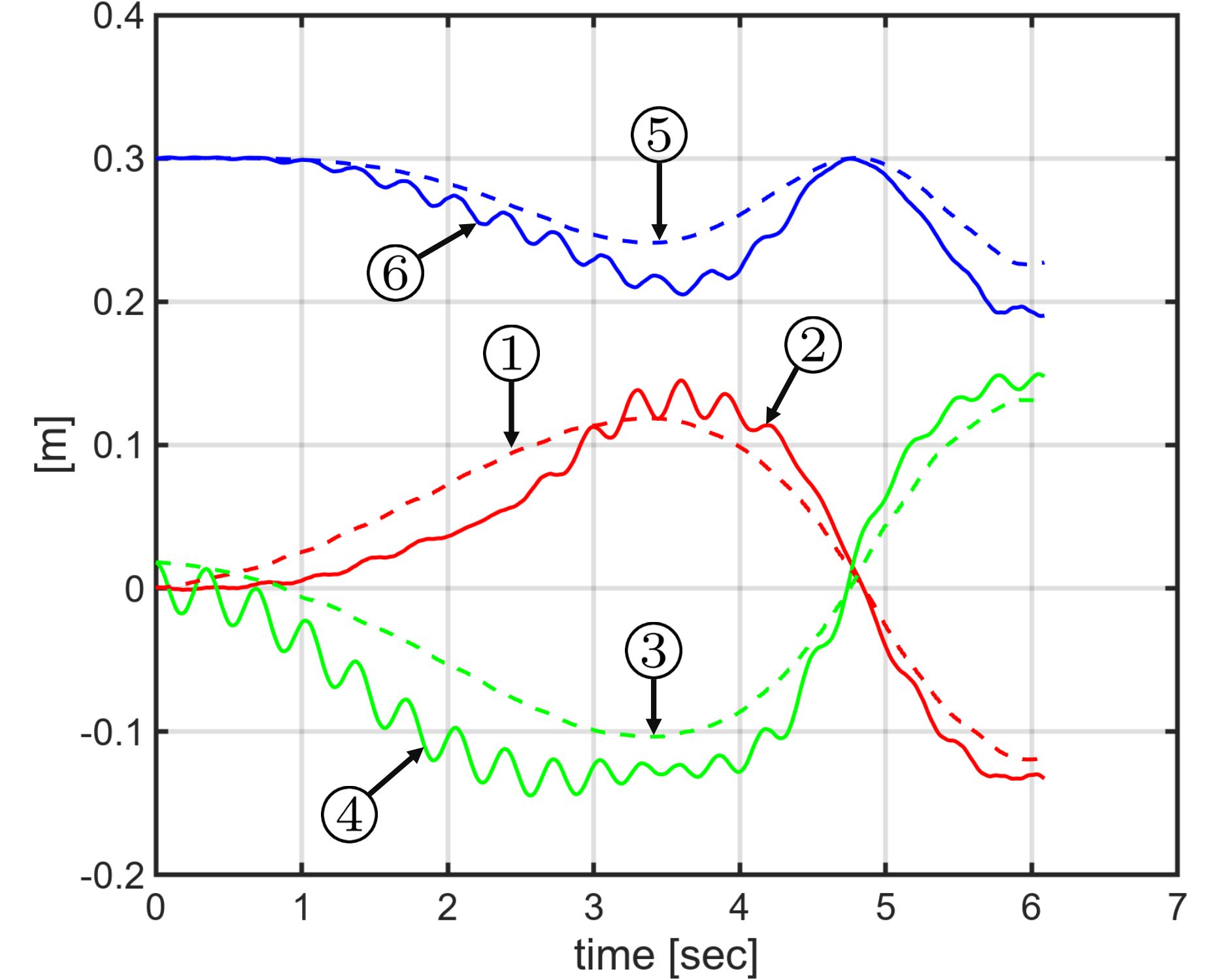}
  \caption{Simulated vs. measured continuum segment tip positions after calibration: \protect\circled{1} Experimental tip $x$ position, \protect\circled{2} simulated tip $x$ position, \protect\circled{3} experimental tip $y$ position, \protect\circled{4} simulated tip $y$ position, \protect\circled{5} experimental tip $z$, \protect\circled{6} simulated tip $z$ position.}\label{fig:calibration_results}
\end{figure}

 Figure \ref{fig:calibration_results} compares the simulated and experimental $x$, $y$, and $z$ components of the segment tip position ${}^0\mb{p}(L)$ using the calibrated parameters. Figure \ref{fig:calibration_results}-\protect\circled{1} shows the experimental tip $x$ position and Fig. \ref{fig:calibration_results}-\protect\circled{2} shows the simulated tip $x$ position. The root-mean-square error (RMSE) in the $x$ direction was 21.7 mm (7.2\% of segment length). Figure \ref{fig:calibration_results}-\protect\circled{3} shows the experimental tip $y$ position and Fig. \ref{fig:calibration_results}-\protect\circled{4} shows the simulated tip $y$ position. The RMSE in the $y$ direction was 35.3 mm (11.7\% of segment length). Figure \ref{fig:calibration_results}-\protect\circled{5} shows the experimental tip $z$ position and Fig. \ref{fig:calibration_results}-\protect\circled{6} shows the simulated tip $z$ position. The RMSE in the $z$ direction was 20.3 mm (6.8\% of segment length).
\par Other than the errors in the dynamic model, potential sources of error in this experiment include the use of commanded current to estimate the applied torque (the actual current supplied to the motor may differ from the command), the lack of exact data synchronization inherent in the decentralized structure of the control framework, \cut{errors in the shape sensing as discussed in \cite{Orekhov2023_shape_sensing}, }errors due to numerical differentiation of the modal coefficients, backlash in the drive train, and sensor noise.  \corrlab{R2-3.2}{Another major source of error was the shape sensing approach first presented in \cite{Orekhov2023_shape_sensing}. In this paper, the authors report an average end-disk positional error of 5.9 mm (1.9\% of $L$) and a maximum error of 14.4 mm (4.8\% of $L$).}

%\begin{figure}[H]
% \centering
%  \includegraphics[width=0.99\columnwidth]{figures/png/chirp_trajectory.png}
%  \caption{Capstan angle excitation trajectory}\label{fig:chirp}
%\end{figure}
\section{Simulation Studies}\label{sec:simulations}
\subsection{Effect of State-Uncertainty}
\begin{figure*}[htbp]
  \centering
  \includegraphics[width=0.9\columnwidth]{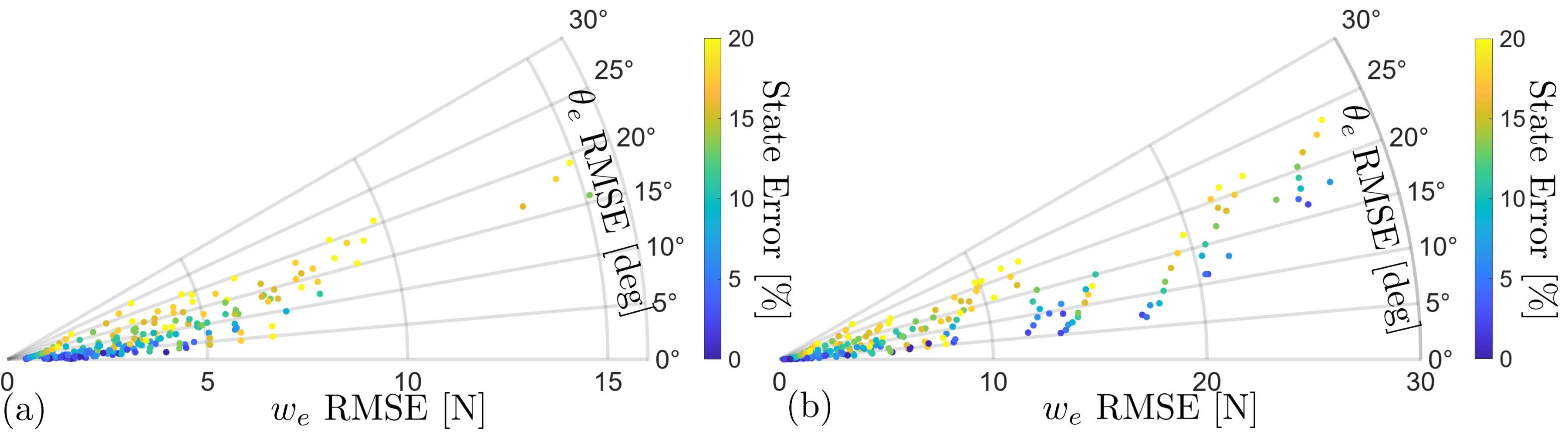}
  \caption{Simulation results: (a) GMO, (b) JFD. In both plots, the radial axis shows the RMSE in the sensed force norm \corrlab{R2-6.3}{$w_e$}, the angular axis shows the RMSE in the estimated wrench direction \corrlab{R2-6.4}{$\theta_e$}, and the color of the data points represent the percent error in the state vector.}\label{fig:sim_results}
\end{figure*}
\par To investigate the effects of state uncertainty on the output of the momentum observer and the joint force/torque deviation method, we generated 21 body frame wrenches with $x$ and $y$ forces ranging between $\pm50$ N. For each wrench, we also generated capstan torque values ranging between $\pm2$ Nm. Using MATLAB's implementation of the Dormand-Prince version of the Runge-Kutta method (\emph{ode45}) (\cite{dormand_prince1980_ode45}), the segment's motion was simulated for 1 second for each randomly generated applied wrench/capstan torques. The simulations were started with the segment in the straight configuration and the applied wrench and capstan torques were increased from zero to the generated value over the 1 second simulation. The wrench was applied to the tip of the continuum robot ($s_c = L$). For each of the 21 generated wrenches, the applied wrench was estimated with constant errors in the state vector ranging from a 0\% to 20\% in steps of 2.22\% for a total of 210 simulation conditions. For the GMO, the observer gain used in the simulation was $\mb{K}_o = 25\mb{I}\in\realfield{6\times6}$, the estimation weight matrix was the identity matrix $\mb{W} = \mb{I}\in\realfield{6\times6}$, and the estimation window was the entire 1 second trajectory. The constraint matrix used in these simulations is given in \eqref{eq:A}. For the JFD method, $\ddot{\mb{c}}$ was estimated using the numerical derivative of the output of dynamic simulation.  
\begin{table*}[ht]
    \centering
    \caption{Simulation results: Mean and max estimated force RMSE values at representative state vector percent errors.}
    \begin{tabular}{@{}cccccc@{}} 
        \toprule
        \textbf{Estimator} & \textbf{state error} & \textbf{mean}($x$) & \textbf{max}($x$) & \textbf{mean}($y$)  & \textbf{max}($y$)  \\ \midrule
       \multirow{4}{*}{GMO} & 0\%     & 1.02 N & 2.88 N & 1.07 N & 2.73 N \\ 
                            & 4.44\%  & 1.50 N & 3.48 N & 1.28 N & 3.14 N\\ 
                            & 11.11\% & 1.98 N & 3.98 N & 2.68 N & 6.91 N \\ 
                            & 15.56\% & 2.52 N & 6.77 N & 3.95 N & 11.60 N  \\ 
                            & 20\%    & 2.96 N & 6.12 N & 5.17 N & 13.56 N \\ \midrule
        \multirow{4}{*}{JFD} & 0\%    & 1.77 N & 5.10 N & 1.30 N & 4.82 N\\ 
                            & 4.44\%  & 4.66 N & 16.11 N & 4.10 N & 20.29 N\\ 
                            & 11.11\% & 4.92 N & 17.00 N & 5.51 N & 20.66 N \\ 
                            & 15.56\% & 5.57 N & 17.55 N & 6.64 N & 21.22 N  \\ 
                            & 20\%    & 5.93 N & 18.46 N & 7.75 N & 22.28 N \\ \midrule
    \end{tabular}
    \label{tab:sim_results}
\end{table*} 
\par The results of these simulations are shown in Fig. \ref{fig:sim_results} and Table \ref{tab:sim_results}. Figure \ref{fig:sim_results} shows polar scatter plots of the simulation results. Figure \ref{fig:sim_results}(a) shows the results for the GMO and Fig. \ref{fig:sim_results}(b) shows the results for the JFD method. In both plots, the radial coordinate indicates the RMSE in the estimated force magnitude \corrlab{R2-6.1}{$w_e = \|\mb{w}_c - \widetilde{\mb{w}}_c\|$} and the angular coordinate indicates the RMSE in the force direction \corrlab{R2-6.2}{$\theta_e = \arccos\left(\mb{w}\T_c \widetilde{\mb{w}}_c\right)/\left(\|\mb{w}_c\| \|\widetilde{\mb{w}}_c\|\right)$}
The color of the dots indicates the percent error in the state vector. 
\par Table \ref{tab:sim_results} shows the mean and max RMSE values at representative percent errors in the state vector. These results indicate that, assuming the applied forces are bounded to $\pm50$ N and the error in the dynamic state is bounded to 20\%, the average RMSE in the GMO force estimation is expected to be around 2.96 N in the $x$ direction and 5.17 N in the $y$ direction. The RMSE can be expected to be below 6.12 N in the $x$ direction and 13.56 N in the $y$ direction. For the JFD method, the average RMSE can be expected to be around 5.93 N in the $x$ direction and 7.75 N in the $y$ direction. The RMSE for the JFD method can be expected to be at or below 18.46 N in the $x$ and 22.28 N in the $y$. 
\par The largest reason for the difference between the errors in the GMO and JFD method is the need for a numerical second derivative of the modal coefficients which introduces additional errors into the system. In these simulations and the experiments in section \ref{sec:experiments}, the estimation errors for forces in the $x$ direction are lower than in the $y$ direction. There are many sources that contribute to this difference. The biggest contributors are the difference between the calibrated flexural rigidities and friction coefficients in Table \ref{tab:calib_results}. Additionally, there is small difference in the elements of $\bs{\mathcal{I}}_{d,i}$ as shown in Table \ref{tab:dynamic_params}.

\subsection{Effect of Sensor Noise}
\begin{figure*}[h]
  \centering
  \includegraphics[width=0.9\columnwidth]{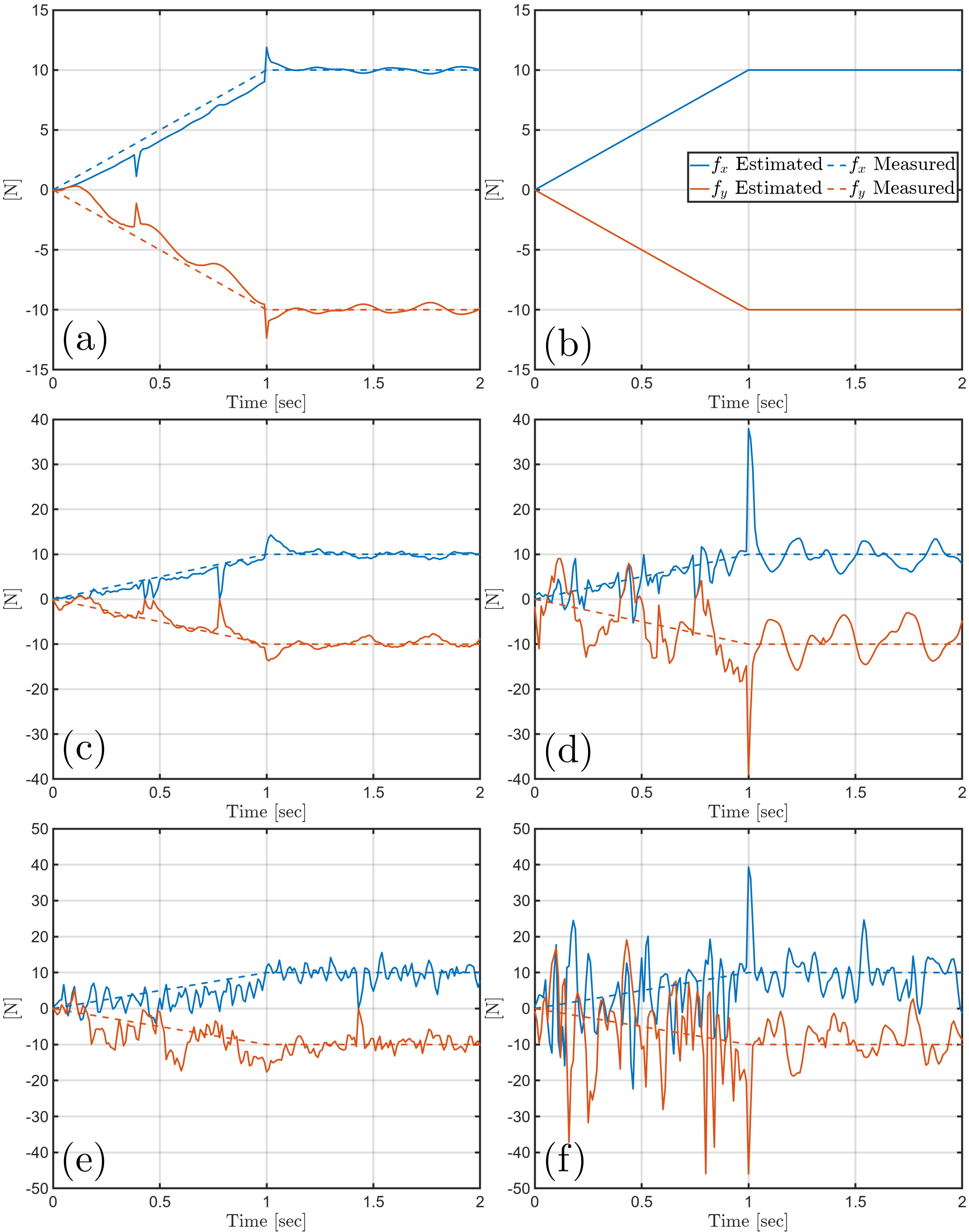}
  \caption{\corrlab{R1-7.2}{Effect of sensor noise simulation results: (a) GMO with no sensor noise, (b) JFD with no sensor noise. (c) GMO with 0.001 peak-to-peak amplitude random noise added to $\mb{c}$. (d) JFD with 0.001 peak-to-peak amplitude random noise added to $\mb{c}$. (e) GMO with 0.01 peak-to-peak amplitude random noise added to $\mb{c}$. (f) JFD with 0.01 peak-to-peak amplitude random noise added to $\mb{c}$.}}\label{fig:noise_sim_results}
\end{figure*}
\begin{table*}[htp]
\caption{\corrlab{R1-7.3}{Sensor noise simulation results for GMO and JFD at increasing noise levels}}\label{tab:noise_sim}
\begin{tabular}{@{}ccccc@{}}
\toprule
\multicolumn{1}{l}{\multirow{2}{*}{\textbf{Method}}} & \multicolumn{1}{l}{\multirow{2}{*}{\textbf{Direction}}} & \multicolumn{3}{c}{\textbf{RMSE {[}N{]}}} \\ \cmidrule(l){3-5} 
\multicolumn{1}{l}{} & \multicolumn{1}{l}{} & Noiseless & 0.001 Amp Noise & 0.01 Amp Noise \\ \midrule
\multirow{2}{*}{GMO} & $x$ & 0.67 & 1.35 & 3.20 \\
 & $y$ & 0.78 & 1.44 & 3.59 \\ \midrule
\multirow{2}{*}{JFD} & $x$ & 9.57$\times10^{-7}$ & 3.89 & 7.71 \\
 & $y$ & 1.26$\times10^{-6}$ & 5.37 & 9.43 \\ \bottomrule
\end{tabular}
\end{table*}

\corrlab{R1-7.4}{As previously stated, a major advantage of the GMO is the ability to filter sensor noise. In this section, we present a simple simulation study showing how sensor noise affects the performance of the GMO and JFD methods. In this simulation study, we simulated the dynamics of the robot for a two second interval. During the first second, $\mb{w}_c$ was ramped up from zero to $[10,-10,0,0,0,0]\T$ and then held steady for the rest of the simulation. The noiseless values of $\mb{c}$, $\dot{\mb{c}}$, and $\ddot{\mb{c}}$ were then plugged directly into the GMO and JFD estimators. In these simulations, the contact location was the robot tip ($s_c = L$), no joint torque was applied to the robot $\bs{\tau} = [0,0]\T$ Nm, and the GMO estimation gain was set to $\mb{K}_O = 10\mb{I}$. No filtering was applied to the observer outputs or wrench estimates. The estimated wrenches for this noiseless simulation are given for the GMO and JFD  in Fig. \ref{fig:noise_sim_results}(a) and \ref{fig:noise_sim_results}(b), respectively.
\par Random noise with 0.001 peak-to-peak amplitude was introduced to $\mb{c}$. Using this noisy version of $\mb{c}$, $\dot{\mb{c}}$ and $\ddot{\mb{c}}$ were estimated using numerical differentiation. For this numerical differentiation, a 10-point Gaussian filter was used to smooth $\mb{c}$ and $\dot{\mb{c}}$ before differentiating. Figure \ref{fig:noise_sim_results}(c) shows the result of the GMO estimation with this noisy state vector, and Fig. \ref{fig:noise_sim_results}(d) shows the results for the JFD. Lastly, this same process was repeated with 0.01 peak-to-peak amplitude random noise added to $\mb{c}$. The results for the GMO and JFD are given in Fig. \ref{fig:noise_sim_results}(e) and Fig. \ref{fig:noise_sim_results}(f), respectively. The RMSE for each noise level is given in Table \ref{tab:noise_sim}.
\par As can be seen in Fig. \ref{fig:noise_sim_results} and Table \ref{tab:noise_sim}, when the state vector and dynamic parameters are known exactly, the JFD perfectly predicts the applied wrench to within numerical precision. On the other hand, the GMO has small differences in the estimation mostly due to the sensing delay and estimator dynamics as discussed in section \ref{sec:estimator_dynamics}. However, when noise is introduced into the simulation, the GMO becomes more accurate for two primary reasons: 1) as discussed in section \ref{sec:estimator_dynamics} the GMO acts as a low-pass filter, and 2) the GMO does not require the numerical second derivatives of $\mb{c}$.}
\section{Experimental Evaluation}\label{sec:experiments}
\begin{figure*}[htbp]
  \centering
  \includegraphics[width=0.99\columnwidth]{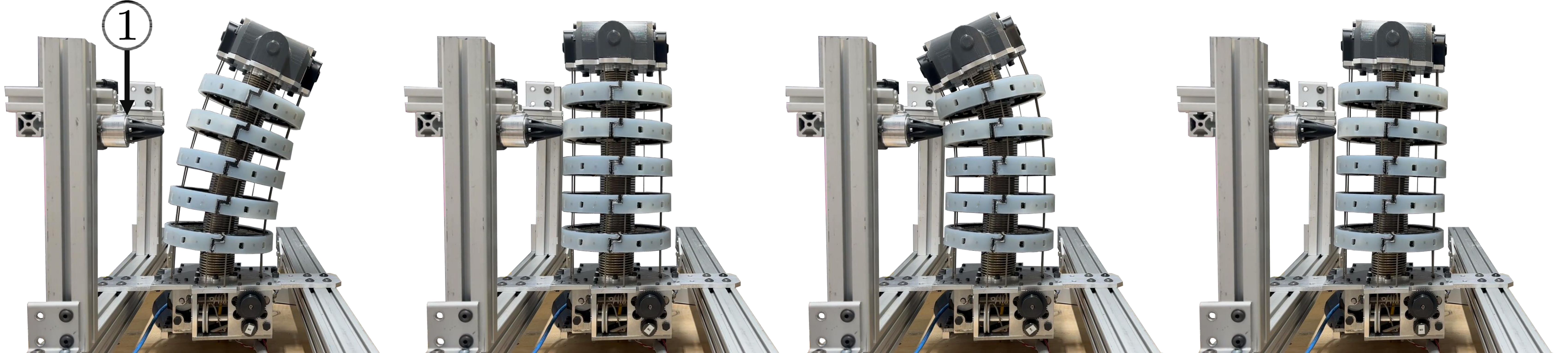}
  \caption{Film strip showing an example of the experimental results given in Table \ref{tab:exp_results}: \protect\circled{1} Bota Systems\texttrademark~Rokubi force/torque sensor.}\label{fig:exp_setup}
\end{figure*}
\begin{figure*}[htbp]
  \centering
  \includegraphics[width=0.95\textwidth]{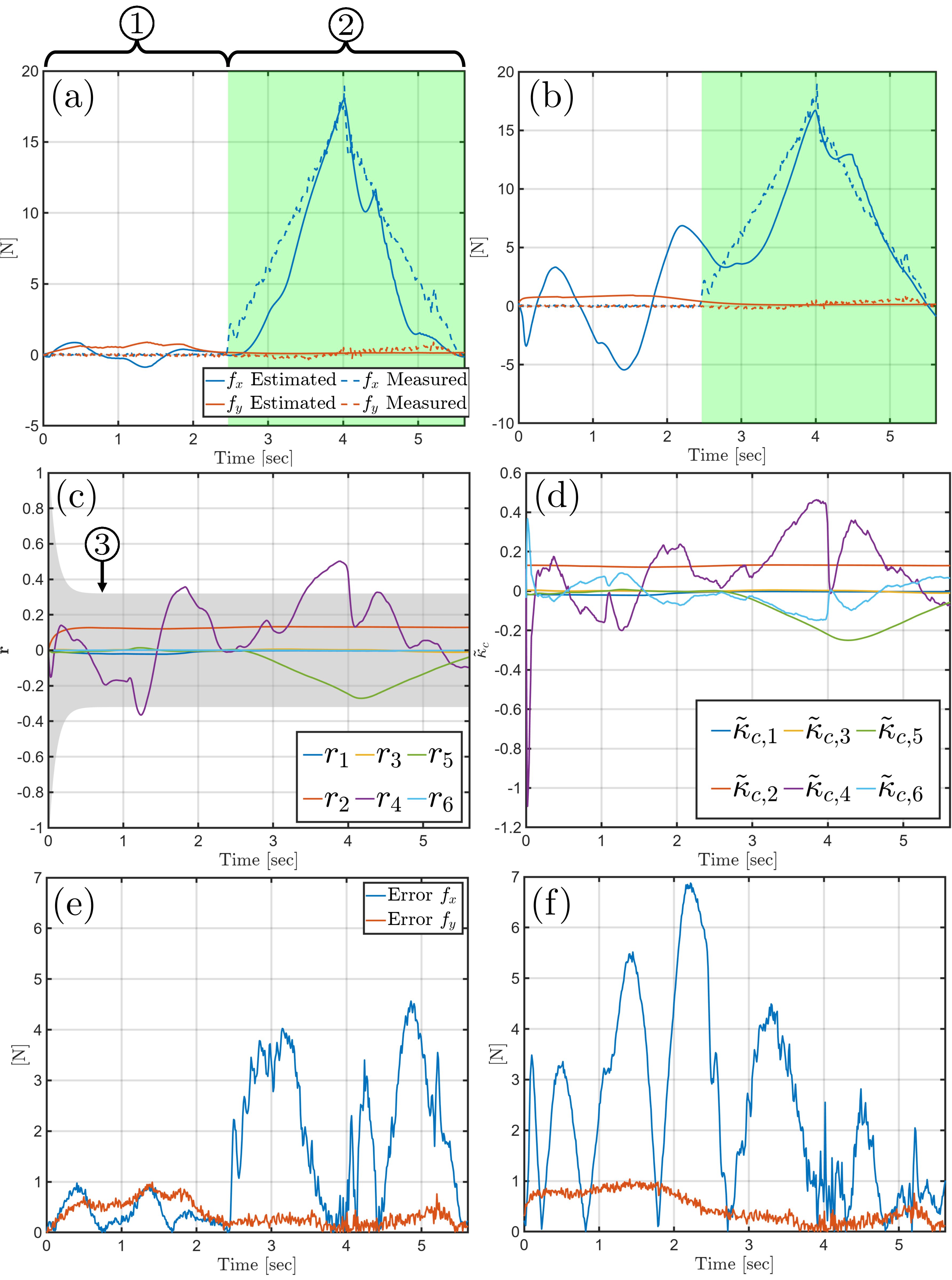}
  \caption{Plots showing an example of the experimental results given in Table \ref{tab:exp_results}. This is the same experiment shown in Fig. \ref{fig:exp_setup}. (a) The measured vs. GMO-estimated applied force in the local disk frame. (b) The measured vs. JFD estimated applied force in the local disk frame. (c) The GMO-estimated generalized forces $\mb{r}$. (d) The JFD-estimated generalized forces $\widetilde{\bs{\kappa}}_c$. \corrlab{R3-2.1}{(e) The absolute value of the error in the local $x$ and $y$ directions for the GMO. (f) The absolute value of the error in the local $x$ and $y$ directions for the JFD. }\protect\circled{1} time span without contact, \protect\circled{2} time span with contact, \protect\circled{3} threshold on $\mb{r}$ above which the generalized force is assumed to be from external contact.}\label{fig:exp_results}
\end{figure*}
\begin{table*}[]
\caption{Experimental test cases and results. The highlighted column indicates the experiment that was plotted in Fig. \ref{fig:exp_results}.}\label{tab:exp_results}
\begin{tabular}{|c|c|ccccc|ccccc|}
\hline
\multicolumn{2}{|c|}{Direction}               & $+x$ & $+x$  & $+x$  & \cellcolor[gray]{0.85}$+x$  & $+x$  & $+y$  & $+y$  & $+y$  & $+y$  & $+y$ \\
\multicolumn{2}{|c|}{Disk Number}             & 1    & 2     & 3     & \cellcolor[gray]{0.85}4     & 5     & 1     & 2     & 3     & 4     & 5 \\
\multicolumn{2}{|c|}{$s_c$ {[}mm{]}}          & 58.5 & 110.0 & 161.8 & \cellcolor[gray]{0.85}213.3 & 265.0 & 58.5  & 110.0 & 161.8 & 213.3 & 265.0 \\
\multicolumn{2}{|c|}{$\dot{q}_1$ {[}deg/s{]}} & 36   & 72    & 108   & \cellcolor[gray]{0.85}144   & 180   & 0     & 0     & 0     & 0     & 0 \\
\multicolumn{2}{|c|}{$\dot{q}_2$ {[}deg/s{]}} & 0    & 0     & 0     & \cellcolor[gray]{0.85}0     & 0     & -36   & -72   & -108  & -144  & -180 \\ \hline
\multirow{2}{*}{GMO} & RMSE$_x$ {[}N{]}       & 4.62 & 5.67  & 3.18  & \cellcolor[gray]{0.85}1.91  & 3.81  & 0.23  & 0.54  & 0.45  & 0.82  & 1.02 \\
                     & RMSE$_y$ {[}N{]}       & 0.60 & 1.46  & 0.97  & \cellcolor[gray]{0.85}0.42  & 0.82  & 11.78 & 8.84  & 12.67 & 11.45 & 9.56 \\ \hline
\multirow{2}{*}{JFD} & RMSE$_x$ {[}N{]}       & 4.60 & 3.64  & 2.85  & \cellcolor[gray]{0.85}2.97  & 5.13  & 0.23  & 0.54  & 0.45  & 0.94  & 1.08 \\
                     & RMSE$_y$ {[}N{]}       & 0.61 & 1.52  & 1.04  & \cellcolor[gray]{0.85}0.56  & 0.84  & 11.78 & 8.94  & 12.35 & 11.44 & 9.57 \\ \hline \hline

\multicolumn{2}{|c|}{Direction}               & $-x$ & $-x$  & $-x$  & $-x$  & $-x$  & $-y$  & $-y$  & $-y$  & $-y$  & $-y$ \\
\multicolumn{2}{|c|}{Disk Number}             & 1    & 2     & 3     & 4     & 5     & 1     & 2     & 3     & 4     & 5 \\
\multicolumn{2}{|c|}{$s_c$ {[}mm{]}}          & 58.5 & 110.0 & 161.8 & 213.3 & 265.0 & 58.5  & 110.0 & 161.8 & 213.3 & 265.0 \\
\multicolumn{2}{|c|}{$\dot{q}_1$ {[}deg/s{]}} & -180 & -144  & -108  & -72   & -36   & 0     & 0     & 0     & 0     & 0 \\
\multicolumn{2}{|c|}{$\dot{q}_2$ {[}deg/s{]}} & 0    & 0     & 0     & 0     & 0     & 180   & 144   & 108   & 72    & 36 \\ \hline
\multirow{2}{*}{GMO} & RMSE$_x$ {[}N{]}       & 3.98 & 6.63  & 3.97  & 4.37  & 4.68  & 1.39  & 1.13  & 1.29  & 1.34  & 0.78 \\
                     & RMSE$_y$ {[}N{]}       & 0.36 & 0.58  & 1.34  & 1.61  & 1.44  & 14.81 & 9.80  & 9.25  & 8.55  & 5.43 \\ \hline
\multirow{2}{*}{JFD} & RMSE$_x$ {[}N{]}       & 3.97 & 6.48  & 2.77  & 12.89 & 9.34  & 1.39  & 1.12  & 1.12  & 1.35  & 0.84 \\
                     & RMSE$_y$ {[}N{]}       & 0.60 & 0.58  & 1.32  & 1.46  & 1.44  & 14.80 & 9.60  & 9.89  & 7.91  & 5.30 \\ \hline
\end{tabular}
\end{table*}
\begin{table*}[]
\caption{Statistics that summarize the experimental results presented in Tab. \ref{tab:exp_results}}\label{tab:exp_results_summary}
\begin{tabular}{@{}clcccccc@{}}
\toprule
\textbf{Estimator}   & \textbf{Statistic}       & \textbf{Disk 1} & \textbf{Disk 2} & \textbf{Disk 3} & \textbf{Disk 4} & \textbf{Disk 5} & \textbf{All Disks} \\ \midrule
\multirow{4}{*}{GMO} & Average RMSE $x$ {[}N{]} & 2.56   & 3.49   & 2.22   & 2.11   & 2.57   & 2.59      \\
                     & Max. RMSE $x$ {[}N{]}    & 4.62   & 6.63   & 3.97   & 4.37   & 4.68   & 6.63      \\
                     & Average RMSE $y$ {[}N{]} & 6.89   & 5.17   & 6.06   & 5.51   & 4.31   & 5.59      \\
                     & Max. RMSE $y$ {[}N{]}    & 14.81  & 9.80   & 12.67  & 11.45  & 9.56   & 14.81     \\ \midrule
\multirow{4}{*}{JFD} & Average RMSE $x$ {[}N{]} & 2.55   & 2.95   & 1.80   & 4.54   & 4.10   & 3.19      \\
                     & Max. RMSE $x$ {[}N{]}    & 4.60   & 6.48   & 2.85   & 12.89  & 9.34   & 12.89     \\
                     & Average RMSE $y$ {[}N{]} & 6.89   & 5.16   & 6.15   & 5.39   & 4.28   & 5.57      \\
                     & Max. RMSE $y$ {[}N{]}    & 14.80  & 9.60   & 12.35  & 11.44  & 9.57   & 14.80     \\ \bottomrule
\end{tabular}
\end{table*}
\par In order to experimentally validate our method \corrlab{R1-6}{and investigate how the speed, location, and direction of contact affects the results}, we mounted a Bota Systems\texttrademark~Rokubi force/torque sensor (Fig. \ref{fig:exp_setup}-\circled{1}) in a known location relative to the continuum segment such that the positive $z$ axis of the force/torque sensor aligned with the positive $x$ axis of the segment base frame $\{T(0)\}$. The force/torque sensor was positioned such that the tip was barely touching the center of first spacer disk. The segment was then commanded to move away from the sensor with capstan velocities $\dot{\mb{q}} = [0,-180]~^\circ/s$ until $\mb{q}= [0,-180]~^\circ$. The segment was then commanded to move towards the force/torque sensor with capstan velocities $\dot{\mb{q}} = [0,~36]~^\circ/s$ until $\mb{q}= [0,~225]~^\circ$ or the magnitude of the $z$ axis force/torque sensor reading was $\geq 20$ N. Finally, the segment was commanded to move away from the sensor with capstan velocities $\dot{\mb{q}} = [0,-36]~^\circ/s$. This was repeated for each disk and with the $z$ axis of the sensor aligned with the positive $y$, negative $y$, positive $x$, and negative $x$ axes of $\{T(0)\}$ for a total of twenty experiments. The capstan angular velocities (and therefore contact point velocities) were varied among the test cases. Table \ref{tab:exp_results} summarizes the conditions for each of the twenty test cases. The first row of the table shows the disk number contacted during the experiment, the second row shows the axis of $\{T(0)\}$ that the $z$ axis of the force/torque sensor was aligned with during the experiment, the third row shows contact arc lengths $s_c$, the fourth and fifth rows correspond with the speed of capstans 1 and 2 when the robot is moving towards the force/torque sensor, respectively. The sixth and seventh rows show the RMS errors of the estimated force in the $x$ and $y$ directions of the body frame at the contact point $\{T(s_c)\}$. The force/torque sensor was read using RS422 serial communication and published on a ROS topic. Videos of a few representative experiments are shown in Multimedia Extension I.          
\par In the experiments, the modal coefficients $\mb{c}$ were estimated from the string-encoder and actuation tendon readings using \eqref{eq:J_lc}. $\dot{\mb{c}}$ was estimated from $\mb{c}$ by applying a 30 point, backwards facing Gaussian filter to $\mb{c}$ and then taking a numerical derivative. Similarly, $\ddot{\mb{c}}$ was estimated from $\dot{\mb{c}}$ by applying a 30 point, backwards facing Gaussian filter to $\mb{c}$ and then taking a numerical derivative. For all experiments, the contact was assumed to be a point contact (i.e., using the $\mb{A}$ matrix given in \eqref{eq:A}) and the estimation weight matrix was the identity matrix $\mb{W} = \mb{I}\in\realfield{6\times6}$.
\par For the GMO, the residual gain was $\mb{K}_o = \operatorname{diag}([10,~25,~25,~10,~25,~0.02])$. The value of $\mb{K}_o$ was chosen using trial and error. The lower gain values for the first, fourth, and last entries of $\mb{K}_o$ was selected due to higher levels of noise in their respective modal coefficients during the experiments. Because of noise introduced into the system from numerical differentiation of the modal coefficients, a 100 point, backwards-facing Gaussian filter was applied to both the estimation residual $\mb{r}$ and the estimated generalized forces $\widetilde{\mb{w}}_c$. The RMSE of the estimated forces in the disk frame $x$ and $y$ directions are given in Table \ref{tab:exp_results}. Figure \ref{fig:exp_results} shows the estimated forces $\widetilde{\mb{w}}_c$ and estimator residuals $\mb{r}$ for one of the twenty evaluation experiments. Figure \ref{fig:exp_results}(a) compares the estimated $x$ and $y$ forces in the local frame $\{T(s_c)\}$ to the measured $x$ and $y$ forces in the $\{T(s_c)\}$ frame. Figure \ref{fig:exp_results}(c) shows the estimated generalized forces $\mb{r}$ during the experiment.
\par For the JFD, a 200 point, backwards-facing Gaussian filter was applied to both the estimated generalized forces $\widetilde{\bs{\kappa}}_c$ and the estimated wrench $\widetilde{\mb{w}}_c$. The RMSE of the estimated forces in the disk frame $x$ and $y$ directions are given in Table \ref{tab:exp_results}. Figure \ref{fig:exp_results} shows the estimated forces $\widetilde{\mb{w}}_c$ and estimator residuals $\widetilde{\bs{\kappa}}_c$ for one of the twenty evaluation experiments. Figure \ref{fig:exp_results}(b) compares the estimated $x$ and $y$ forces in the local frame $\{T(s_c)\}$ to the measured $x$ and $y$ forces in the $\{T(s_c)\}$ frame. Figure \ref{fig:exp_results}(d) shows the estimated generalized forces $\mb{r}$ during the experiment.
\par The shaded region of Fig. \ref{fig:exp_results}(c) labeled \circled{3} shows the threshold on the elements of $\mb{r}$ above which the generalized forces are assumed to be from external contact and not from uncertainty in the dynamic state as described in section \ref{sec:uncertainty}. The value of $\Delta\mb{x}$ chosen to establish this threshold was 1.84 for all elements. This was 75\% of the maximum error in the dynamic state during the calibration experiments described in section \ref{sec:calibration}. The use of these thresholds helps guard against false positives, but comes at the cost of delayed detection response and sensitivity. Note that each element of $\mb{r}$ has its own threshold and Fig. \ref{fig:exp_results}(c)-\circled{3} only shows the threshold associated with $r_4$ for visual clarity.
\corrlab{R3-2.2}{Figures \ref{fig:exp_results}(e) and \ref{fig:exp_results}(f) show the absolute value of the error in the $x$ and $y$ directions of the body frame at the contact point $\{T(s_c)\}$ for the GMO and JFD, respectively.}
\par Table \ref{tab:exp_results_summary} summarizes the mean and max RMSEs across all 20 experiments and for each intermediate disk. For the GMO, across all 20 experiments, the average RMSE was 2.59 N in the disk-frame $x$ direction and 5.59 N in the disk-frame $y$ direction. For the JFD method, across all 20 experiments, the average RMSE was 3.19 N in the disk-frame $x$ direction and 5.57 N in the disk-frame $y$ direction. When comparing the performance of the GMO and the JFD method, the GMO performed better on average than the JFD method in the $x$ direction. The GMO and JFD method performed similarly on average in the $y$ direction. It is worth noting that for the JFD method to have a similar performance to the GMO, the filter window on the estimation signals had to be twice as long as the GMO. This is because of the GMO only needs the numerical first derivative of $\mb{c}$ while the JFD method requires the numerical second derivative of $\mb{c}$. The largest sources of error in these experiments was most likely the accuracy of the dynamic model and the numerical differentiation of the modal coefficients. \corrlab{R2-3.3}{Another major source of error is the accuracy of the shape sensing approach as detailed in \cite{Orekhov2023_shape_sensing}.} Other sources of error include that the point of contact was assumed to be fixed relative to the spacer disks. However, during the experiments the contact point did move slightly relative to the disk as can be seen in Fig. \ref{fig:exp_setup}. Additionally, the segment's NiTi backbone has a slight permanent bend in the $yz$-plane. This contributes to the larger errors in $y$ direction forces. Neither method was able to detect contact on the first and second disks. This is because the forces were not able to make a significant enough change in the dynamic state to register as a contact.
\corrlab{R2-7}{When comparing the simulations in Table \ref{tab:sim_results} and the experimental result \ref{tab:exp_results_summary}, it can be seen that the performance gain of the GMO over the JFD was less in experiment than in simulation. This is most likely due to the uncertainty in the dynamic parameters in real experimental scenarios.}
\begin{figure*}[htbp]
  \centering
\includegraphics[width=0.99\columnwidth]{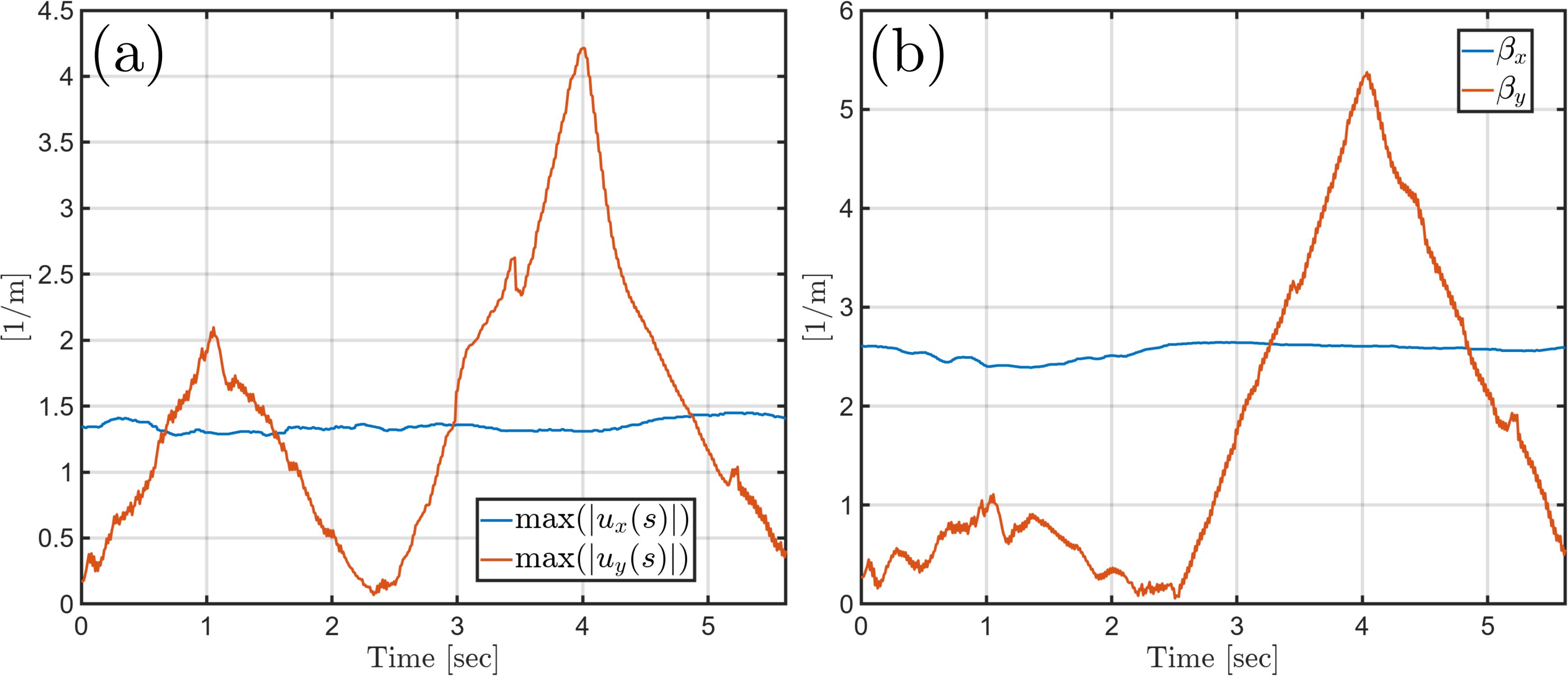}
  \caption{\corrlab{R1-2.2}{(a) Maximum curvature in the local $x$ and $y$ direction during the experiment shown in Fig. \ref{fig:exp_setup}. (b) Circular-ness measure $\beta$ in the local $x$ and $y$ direction during the experiment shown in Fig. \ref{fig:exp_setup}.}}\label{fig:curvature}
\end{figure*}
\par \corrlab{R1-2.1}{The vast majority of prior works on continuum robot contact detection assume a constant curvature (i.e. circular) bending model. A key contribution of this paper is a method for robots that do not bend in constant curvature arcs. To demonstrate that the robot did not bend in constant curvature during the experiment shown in Fig. \ref{fig:exp_setup}, we introduce a simple measure of how non-circular the robot was bending at any given time. First we define, the maximum curvature along the robot's arclength, $u_{\max} = \max(u(s)),~s = [0,L]$, and the minimum curvature along the robot's arclength, $u_{\min} = \min(u(s)),~s = [0,L]$. The measure $\beta$ is simply:
\begin{equation}
    \beta_x = u_{x,\max} - u_{x,\min}, \quad \beta_y = u_{y,\max} - u_{y,\min}  
\end{equation}
The closer $\beta$ is to zero, the better the robot's bending can be approximated by a circular arc. Figure \ref{fig:curvature}(a) shows the maximum curvature during the experiment shown in Fig. \ref{fig:exp_setup}. Figure \ref{fig:curvature}(b) shows the circular-ness measure $\beta$ during the experiment.} 

\section{Preliminary Experimental Extension to Multi-Segment Continuum Robots}\label{sec:multiseg}
\corrlab{R1-1.1}{The results given in the previous section were for the distal segment in isolation. In this section, we will show preliminary results extending the method to the multisegment continuum robot shown in Fig. \ref{fig:Vanderbot}. To do this, we must first discuss how the reaction moments from the distal segment's actuation affects the dynamics of the proximal segment.
\subsection{Reaction Wrenches on the Proximal Segment}
In these experiments, the mass and inertia of the distal segment was statically lumped into the end-disk. Additionally, the proximal segment must resist the reaction moments from the distal segment's actuation motors as well as the reaction moment from the elbow motor placed between the two segments (See Fig. \ref{fig:multiseg_setup}-\circled{4}). To include these moments, we included the generalized forces due to the distal segment, $\bs{\kappa}_{dist}$, into the non-conservative forces given in \eqref{eq:non_cons_gen_force}: 
\begin{equation}
    \bs{\kappa} = \bs{\kappa}_{dist} + \mb{J}\T_{qc}\bs{\tau} + \mb{J}_{\xi c}\T(s_c)\mb{w}_c - \bs{\kappa}_{fric}
\end{equation}
To calculate $\bs{\kappa}_{dist}$, the wrench on the end-disk due to the elbow motor, $\mb{w}_{elbow}$, and from the distal segment actuation motors, $\mb{w}_{\tau,dist}$ are multiplied by the transpose of the proximal segment Jacobian:
\begin{equation}
    \bs{\kappa}_{dist} = \mb{J}_{\xi c}\T(L)\left(\mb{w}_{elbow}+\mb{w}_{\tau,dist}\right)
\end{equation}
In these equations, $\mb{w}_{elbow} = [0,0,0,\tau_{elbow},0,0]\T$ and $\mb{w}_{\tau,dist} = [0,0,0,\tau_{dist,1},\tau_{dist,2},0]$.
\subsection{Proximal Segment Calibration}
Now that the effect of the distal segment's actuation on the proximal segment is accounted for, we can calibrate the proximal segment's dynamic parameters. To do this, a similar process to the one described in section \ref{sec:calibration} was completed. In the calibration experiment, the distal segment was attached to the proximal segment and the mass and inertia of the distal segment were lumped into the end-disk parameters of the proximal segment. The calibrated parameters are given in Table \ref{tab:calib_results_prox}.  
\begin{table}[ht]
    \centering
    \caption{\corrlab{R1-1.2}{Proximal segment calibrated dynamic model parameters}}
    \begin{tabular}{@{}cccc@{}} 
        \toprule
        $EI_x$ & $EI_y$ & $\mu_1$ & $\mu_2$ \\ \midrule
        1.6003 & 2.1502 & 0.0900  & 0.2000 \\ \bottomrule
    \end{tabular}
    \label{tab:calib_results_prox}
\end{table} 
\subsection{Experimental Setup}
\begin{figure*}
    \centering
\includegraphics[width=1\linewidth]{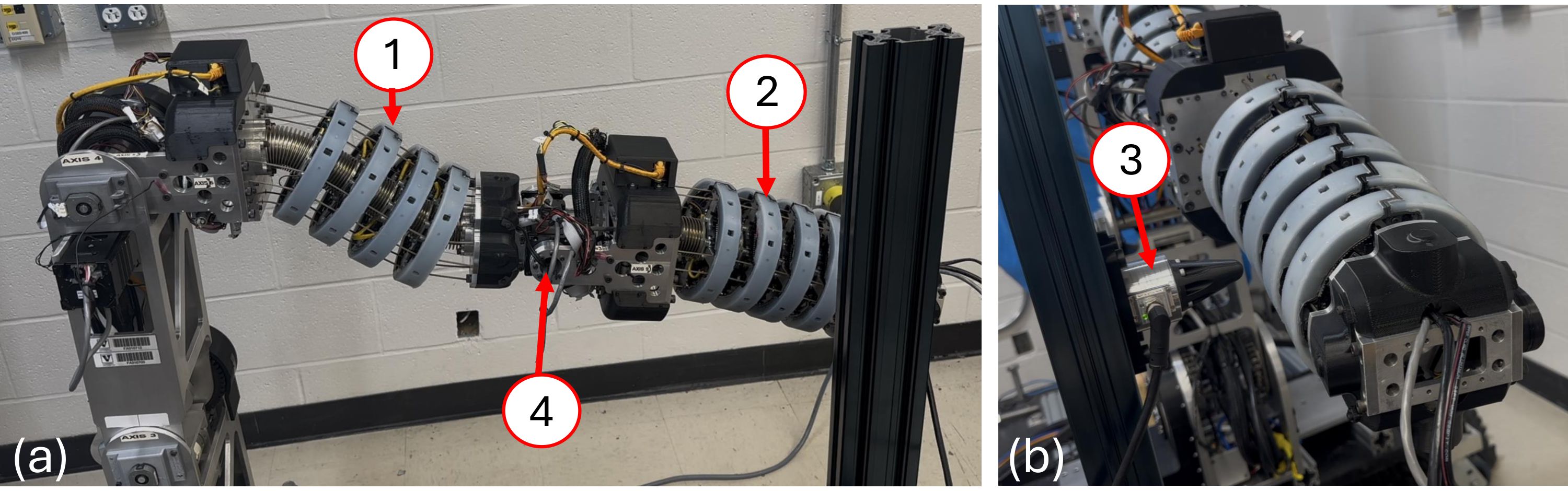}
    \caption{\corrlab{R1-1}{Multisegment experimental setup. (a) Robot at its starting configuration. (b) Close-up of the contact point. \protect\circled{1} Proximal segment, \protect\circled{2} distal segment, \protect\circled{3} Rokubi Force/Torque Sensor, \protect\circled{4} elbow motor between segments. Videos of these experiments are provided in multimedia extension II.}}
    \label{fig:multiseg_setup}
\end{figure*}
In these experiments, we placed the continuum segments in a horizontal configuration as shown in Fig. \ref{fig:multiseg_setup} ($\mb{g} = [0,9.81,0]\T~m/s^2$). Again, the Bota Systems Rokubi force/torque sensor was placed in the workspace of the robot. In this configuration, similar to the single-segment experiments, both segments were commanded to move away from the sensor with capstan velocities $\dot{\mb{q}} = [0,-180]~^\circ/s$ until $\mb{q}= [0,-180]~^\circ$. The segments were then commanded to move towards the force/torque sensor with capstan velocities $\dot{\mb{q}} = [0,~36]~^\circ/s$ until $\mb{q}= [0,~225]~^\circ$ or the magnitude of the $z$ axis force/torque sensor reading was $\geq 30$ N. Finally, the segments were commanded to move away from the sensor with capstan velocities $\dot{\mb{q}} = [0,-36]~^\circ/s$. Videos of these experiments are provided in multimedia extension II. 
\par In the experiments, the robot came into contact with the 4th spacer disk of the distal segment $s_c = 213.293$ mm. The observer gains and filter windows were the same as the single-segment experiment.
\subsection{Results}
\begin{figure*}
    \centering
    \includegraphics[width=1\linewidth]{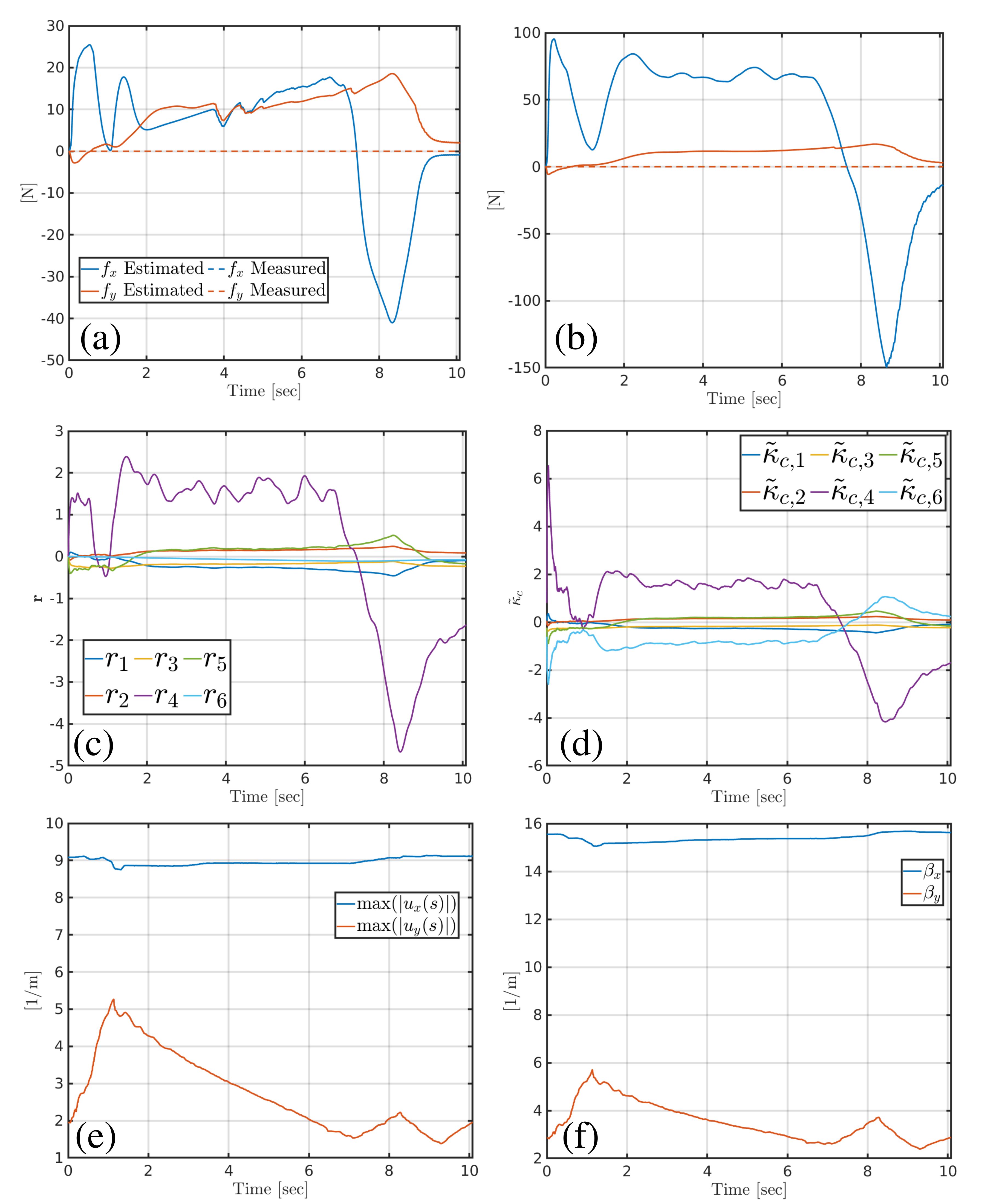}
    \caption{\corrlab{R1-1}{Experimental results for the proximal segment. (a) Estimated forces in the $x$ and $y$ directions using the GMO. (b) $x$ and $y$ directions using the JFD. (c) GMO-estimated generalized forces $\mb{r}$. (d) (d) The JFD-estimated generalized forces $\widetilde{\bs{\kappa}}_c$. (e) The maximum curvature in the local $x$ and $y$ directions. (f) The circular-ness measure $\beta$ in the local $x$ and $y$ directions.}}
    \label{fig:prox_results}
\end{figure*}
\begin{figure*}
    \centering
    \includegraphics[width=1\linewidth]{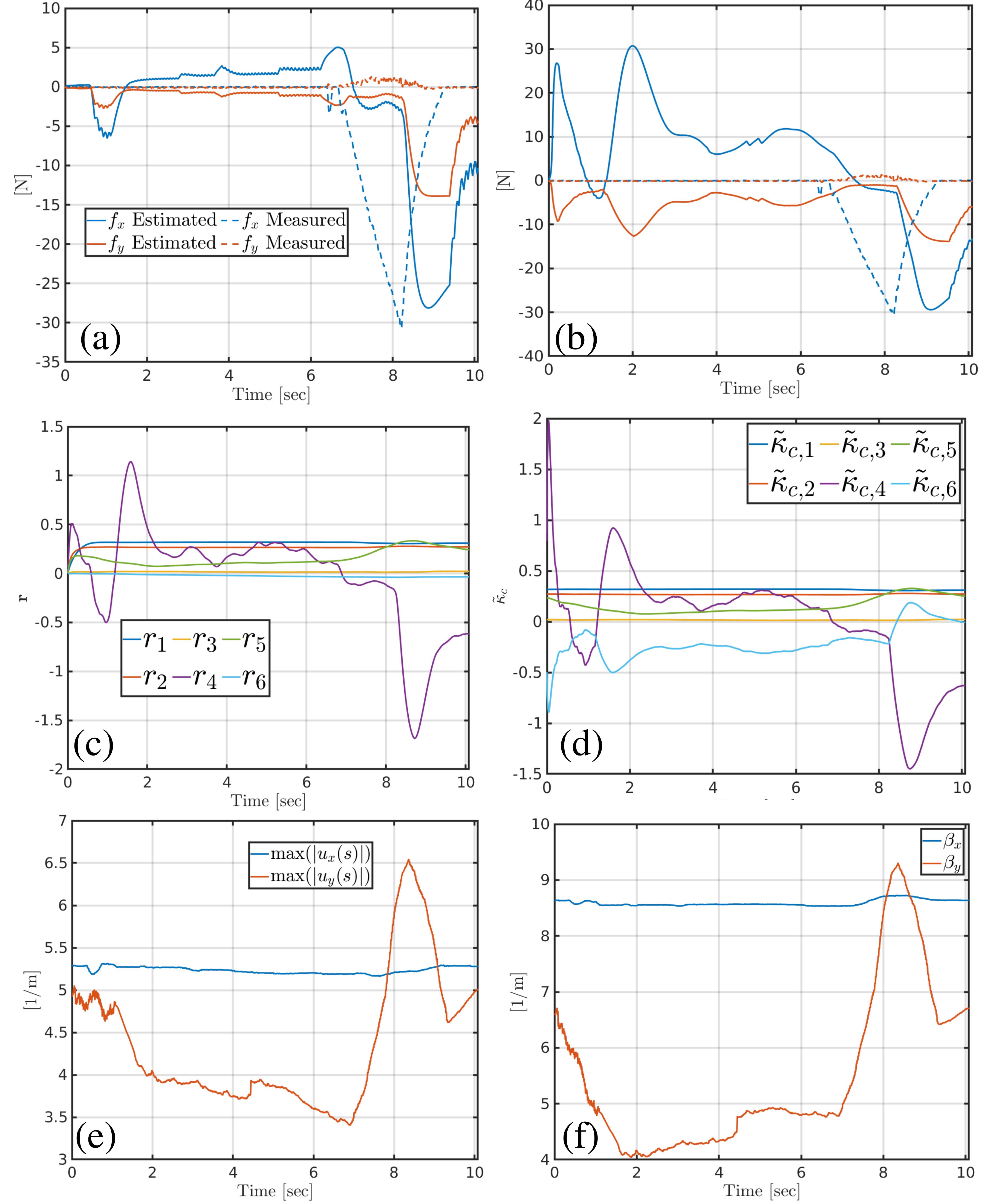}
    \caption{\corrlab{R1-1}{Experimental results for the distal segment. (a) Estimated forces in the $x$ and $y$ directions using the GMO. (b) $x$ and $y$ directions using the JFD. (c) GMO-estimated generalized forces $\mb{r}$. (d) (d) The JFD-estimated generalized forces $\widetilde{\bs{\kappa}}_c$. (e) The maximum curvature in the local $x$ and $y$ directions. (f) The circular-ness measure $\beta$ in the local $x$ and $y$ directions.}}
    \label{fig:dist_results}
\end{figure*}
Figure \ref{fig:prox_results} shows the experimental results for the proximal continuum segment and Fig \ref{fig:dist_results}. shows the results for the distal segment. In both figures, (a) shows the estimated wrench using the GMO, (b) shows the estimated wrench on the proximal segment using the JFD. Subfigures (c) and (d) show the estimated generalized forces due to contact for the GMO and JFD methods, respectively. Lastly, (e) shows the maximum curvature during the experiment and (f) shows the circular-ness measure during the experiment. Table \ref{tab:two_seg} shows the estimation RMSE for the distal and proximal segments using both the JFD and GMO methods.   
\begin{table}[]
\caption{\corrlab{R1-1}{RMSE values for the proximal and distal segments}}\label{tab:two_seg}
\begin{tabular}{@{}cccc@{}}
\toprule
Segment                   & Method               & Direction & RMSE {[}N{]} \\ \midrule
\multirow{4}{*}{Proximal} & \multirow{2}{*}{GMO} & $x$       & 16.07        \\
                          &                      & $y$       & 10.51        \\ \cmidrule(l){2-4} 
                          & \multirow{2}{*}{JFD} & $x$       & 68.01        \\
                          &                      & $y$       & 10.74        \\ \midrule
\multirow{4}{*}{Distal}   & \multirow{2}{*}{GMO} & $x$       & 10.45        \\
                          &                      & $y$       & 4.65         \\ \cmidrule(l){2-4} 
                          & \multirow{2}{*}{JFD} & $x$       & 15.96        \\
                          &                      & $y$       & 6.55         \\ \bottomrule
\end{tabular}
\end{table}
\subsection{Discussion and Limitations}
\par As can be seen in Fig. \ref{fig:multiseg_setup}, the proximal segment is bent in a noticeable S-curve shape. This shape could not be reasonable approximated using a constant curvature assumption. 
In these experiments, both methods performed worse during the multisegment experiments than the single segment experiments. This is mostly because the segments GMO and JFD estimators were mostly run in isolation with minimal coupling effects. For example, the kinetic energy due to the motion in the proximal segment was not taken into account for the distal segments and the reaction forces due to motion of the distal segments were not taken into account for the proximal segment. Additionally, the distal segment was approximated as a fixed mass and inertia and lumped into the end-disk parameters. This approximation was a major source of error.
\par Also, although there was not contact with the proximal segment, the GMO detected contact as can be seen in Fig. \ref{fig:prox_results}(a). This is due to the contact on the distal segment. The GMO was therefore estimating an equivalent wrench at its fourth disk that would cause the same change in dynamics as the true contact with the distal segment. In practice, the contact sensors in the robot's sensor disk (see \cite{Abah2022_sensor_disk}) could be used to determine which portion of the robot was in contact and the GMO could be used to determine the location of contact. 

}
\section{Conclusions}
To safely operate alongside workers in a confined space, \textit{in-situ} collaborative robots (ISCRs) must be endowed with a host of sensory information not typically present in traditional industrial robots. One of the most important of these sensory modalities is the ability to detect and estimate wrenches applied to the robot by the environment or by the human operator.    
\par In this paper, we presented a formulation of the generalized momentum observer (GMO) in the space of modal coefficients of curvature. This formulation enables contact estimation for continuum robots with variable curvature. We presented a modal-space dynamic model for the continuum segment shown in Fig. \ref{fig:exp_setup} and calibrated the central backbone flexural-rigidities and static friction coefficients that minimize errors in the dynamic model. Additionally, we presented a constrained minimization approach for estimating the wrench applied to the segment during the contact. This approach allows for incorporating knowledge of the contact type. The effect of dynamic state uncertainty on the estimation signal was then investigated and recommendations were made about establishing contact detection thresholds. A simulation study was presented that establishes limits on the expected performance of the method for up to 20\% error in the segment's dynamic state and compared the performance of the GMO to the joint force/torque deviation (JFD) method. The GMO and JFD methods were also compared using 20 different experimental conditions. In these experiments, the JFD was found to have similar performance in the local $y$ direction (which had worse performance in general), but the GMO was found to have a better performance in the local $x$ direction. \corrlab{R1-1}{We also demonstrated a basic extension of the method for  multisegment continuum robots and showed preliminary experimental results.}
\par A major limitation of this work is that the contact location cannot be determined. In order to achieve this, this method could be used in conjunction with the screw-deviation method presented in \cite{Bajo2012TRO_continuum_collision}. This would have the added benefit of increasing sensitivity of the method at very low speeds. Another limitation of our work is that it does not work effectively for the first and second spacer disks. 
\par Future work will include investigating contact localization, improving the dynamic model by including hysteretic elastic effects, and viscous friction effects. Although our method is the first to be specifically tailored to general curvature robots with significant inertial effects, we will work on a systematic comparison to additional existing contact detection methods to establish a baseline of comparison. Additionally, we will work on real-time non MATLAB-based implementation of this method to enable real-time contact estimation. \corrlab{R1-4.3}{Lastly, we will also investigate how the choice of Chebyshev polynomial order affects the performance of the method.} \cut{Lastly, we will apply the GMO method to the entire robot shown in Fig. \ref{fig:Vanderbot} instead of an isolated continuum segment.}     

% \begin{acks}
% The authors thank Dr. Giuseppe Del Giudice and Neel Shihora for their support in the early stages of this work.
% \end{acks}
\begin{funding}
\par This work was supported in part by the National Science Foundation grant \#1734461 and by Vanderbilt University internal funds.
\end{funding} %

% \begin{dci}
% \par Nabil Simaan has surgical robotics technologies licensed to Intuitive Surgical, Johnson \& Johnson, and Titan Medical.
% \end{dci} 

\appendix
\section*{Dynamic Matrices Derivation}
In this appendix, we expand upon the derivations of the terms in the dynamic model described in section \ref{sec:dynamics}. First, we will expand \eqref{eq:lagrange_first_def}: 
\begin{equation}
    \frac{d}{dt}\frac{\partial T}{\partial \dot{\mb{c}}} - \frac{d}{dt}\cancelto{0}{\frac{\partial V}{\partial \dot{\mb{c}}}}
    -\frac{\partial T}{\partial \mb{c}} + \frac{\partial V}{\partial \mb{c}} = \bs{\kappa} 
\end{equation}
The potential energy does not depend $\dot{\mb{c}}$ so therefore the term $\frac{d}{dt}\frac{\partial V}{\partial \dot{\mb{c}}}$ goes to $\mb{0}$. The partial derivative of the kinetic energy can be written as: 
\begin{equation}
    \frac{\partial T}{\partial \dot{\mb{c}}} = \mb{M}\dot{\mb{c}}
\end{equation}
The time derivative of $\frac{\partial T}{\partial \dot{\mb{c}}}$ can be written as:
\begin{equation}\label{eq:EL_first_term}
    \frac{d}{dt}\frac{\partial T}{\partial \dot{\mb{c}}} = \dot{\mb{M}}\dot{\mb{c}} + \mb{M}\ddot{\mb{c}}
\end{equation}
The $i^\text{th}$ row of \eqref{eq:EL_first_term} can be written as:
\begin{equation}
    \frac{d}{dt}\frac{\partial T}{\partial \dot{c}_i} = \sum_{j=1}^6 M_{ij}\ddot{c_j} + \sum_{j=1}^6\sum_{k=1}^6\frac{\partial M_{ij}}{\partial c_k}\dot{c}_j\dot{c}_k 
\end{equation}
The partial derivative of $\mb{M}$ with respect to the $i^\text{th}$ modal coefficient can be found by differentiating the mass matrix as defined in \eqref{eq:M}:
\begin{equation}
    \frac{\partial \mb{M}}{\partial c_i} = \frac{\partial \mb{M}_{CB}}{\partial c_i} + \frac{\partial \mb{M}_{SD}}{\partial c_i} + \cancelto{0}{\frac{\partial \mb{M}_A}{\partial c_i}} 
\end{equation}
Because $\mb{M}_A$ is a constant matrix and not a function of the modal coefficients, $\frac{\partial \mb{M}_A}{\partial c_i}$ goes to $\mb{0}$. In our implementation, the partial derivatives of $\mb{M}_{CB}$ and $\mb{M}_{SD}$ are found using a central finite difference approximation. 
\par The third term in \eqref{eq:lagrange_first_def}, the partial derivative of the kinetic energy with respect to the $i\text{th}$ modal coefficient, can be written as:
\begin{equation}
    \frac{\partial T}{\partial c_i} = \sum_{j=1}^6\sum_{k=1}^6\frac{1}{2}\frac{\partial M_{jk}}{\partial c_i}\dot{c}_j\dot{c}_k
\end{equation}
\par Next, we will find the partial derivative of the potential energy $V$ with respect to the modal coefficients:
\begin{equation}
    \frac{\partial V}{\partial \mb{c}} = \frac{dV_B}{d\mb{c}} + \frac{dV_{CB}}{d\mb{c}} + \frac{dV_{SD}}{d\mb{c}}
\end{equation}
The derivative of the bending potential energy:
\begin{equation}
    \frac{dV_B}{d\mb{c}} = \frac{1}{2}\left(\mb{K}_c+\mb{K}_c\T\right)\mb{c}
\end{equation}
The derivative of the gravitational potential energy of the central backbone:
\begin{equation}
    \frac{dV_{CB}}{d\mb{c}} = -\rho\int_0^L \left(\frac{\partial\mb{p}(s)}{\partial \mb{c}}\right)\T\mb{g}ds
\end{equation}
The derivative of the gravitational potential energy of the spacer disks:
\begin{equation}
    \frac{dV_{SD}}{d\mb{c}} = -\sum_{i=1}^{n_d} m_{d_i}\left(\frac{d\mb{p}(s_{d_i})}{d\mb{c}}+\frac{d}{d \mb{c}}\Big( \Rot{0}{t}(s_{d_i})\Big)\,{}^t\mb{p}_{cm_i}\right)\T\mb{g}
\end{equation}
\par Combining the above terms, the $i^\text{th}$ row of the segment dynamics can be written as: 
\begin{equation}
    \sum_{j=1}^6 M_{ij}\ddot{c_j} + \sum_{j=1}^6\sum_{k=1}^6\left(\frac{\partial M_{ij}}{\partial c_k} - \frac{1}{2}\frac{\partial M_{jk}}{\partial c_i}\right)\dot{c}_j\dot{c}_k + \frac{\partial V}{\partial c_i} = \kappa_i
\end{equation}
Using Christoffel symbols, the second term in the above equation can be written in terms of the centrifugal/Coriolis matrix, $\mb{N}$. The $ij^\text{th}$ element of $\mb{N}$ is:
\begin{equation}
  N_{ij} = \frac{1}{2}\sum_{k=1}^6\left(\frac{\partial M_{ij}}{\partial c_k} + \frac{\partial M_{ik}}{\partial c_j} - \frac{\partial M_{kj}}{\partial c_i}\right)\dot{c}_k
\end{equation}
\bibliographystyle{SageH}
\bibliography{bib/Garrison,bib/Nabil,bib/continuum_robot_dynamics}
\end{document}